\documentclass[journal]{IEEEtran}

\ifCLASSINFOpdf
\else
   \usepackage[dvips]{graphicx}
\fi
\usepackage{url}

\usepackage{graphicx}
\usepackage{epsfig,epstopdf}
\usepackage{algpseudocode}
\usepackage{balance}
\usepackage{amsmath,amssymb,amsfonts}
\usepackage{multicol,multirow}
\usepackage{mathrsfs}
\usepackage{eufrak}
\usepackage{color, graphics,graphicx}
\usepackage{algorithm}
\usepackage{threeparttable}
\usepackage{subfigure,caption}
\usepackage{cancel}
\usepackage{enumitem}
\usepackage{booktabs}
\newtheorem{theorem}{\bf Theorem}

\def\ie{\emph{i.e., }}
\def\eg{\emph{e.g., }}

\def\yifan{\textcolor{black}}
\def\normaljudy{\textcolor{black}}

\def\ming{\textcolor{black}}
\definecolor{netone}{RGB}{34,117,220}

\definecolor{nettwo}{RGB}{0,130,75}

\definecolor{qiu}{RGB}{250,10,10}

\begin{document}

\title{Collaborative Unsupervised  Domain Adaptation \\for Medical Image Diagnosis}

\author{Yifan Zhang, Ying Wei, Qingyao Wu, Peilin Zhao, Shuaicheng Niu, Junzhou Huang, Mingkui Tan
\thanks{This work was partially supported by Key-Area Research and Development Program of Guangdong Province (2017B090901008, 2018B010107001, 2018B010108002),
National Natural Science Foundation of China (NSFC) 61836003 (key project), 61876208, Guangdong Project  2017ZT07X183,
Fundamental Research Funds for the Central Universities D2191240, Pearl River S$\&$T Nova Program of Guangzhou 201806010081.  \textit{(Y. Zhang, Y. Wei and Q.~Wu contributed equally)(Corresponding to J. Huang
and M. Tan).}}
\thanks{Y. Zhang,  S. Niu are with  South China University of Technology, and Pazhou Lab. E-mail: $\{$sezyifan, sensc$\}$@mail.scut.edu.cn.}
\thanks{Y. Wei, P. Zhao and J. Huang are with Tencent AI Lab, China. Email: $\{$judywei, masonzhao, joehhuang$\}$@tencent.com.}
\thanks{Q. Wu and M. Tan are with South China University of Technology, China. E-mail: $\{$qyw, mingkuitan$\}$@scut.edu.cn.}

}
\markboth{IEEE Transactions on Image Processing}
{Shell \MakeLowercase{\textit{et al.}}: Bare Demo of IEEEtran.cls for IEEE Journals}
\maketitle

\begin{abstract}
Deep learning based medical image diagnosis has shown great potential in clinical medicine. However, it often suffers two major difficulties in real-world applications: 1) only limited labels are available for model training, due to expensive annotation costs over medical images; 2) labeled images may contain  considerable label noise (\eg mislabeling labels) due to diagnostic difficulties of diseases. To address these, we seek to exploit rich labeled data from relevant domains to help the learning in the target task via  {Unsupervised Domain Adaptation} (UDA).  Unlike most UDA methods
that rely on clean labeled data or assume samples are equally transferable, we innovatively propose a Collaborative Unsupervised Domain Adaptation algorithm, which conducts transferability-aware adaptation and conquers label noise in a collaborative way. We theoretically analyze the generalization performance of the proposed method, and also empirically evaluate it on both medical and general images. Promising experimental results demonstrate the superiority and generalization of the proposed method.
\end{abstract}


\IEEEpeerreviewmaketitle

\section{Introduction}
\label{sec:intro}

\begin{figure}[t]
 \begin{minipage}{0.3\linewidth}
 \centerline{\includegraphics[width=2.8cm]{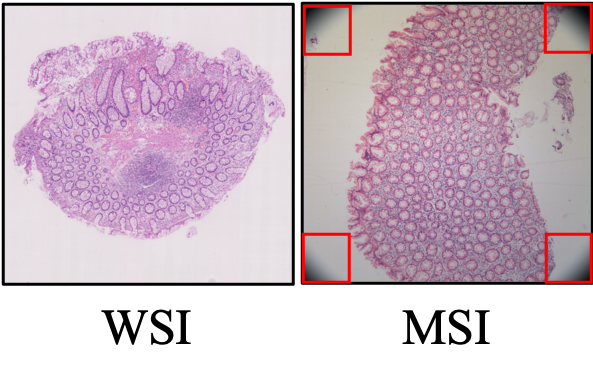}}
 \centerline{(a) scope shades}
 \end{minipage}
 \hfill
  \begin{minipage}{0.3\linewidth}
 \centerline{\includegraphics[width=2.8cm]{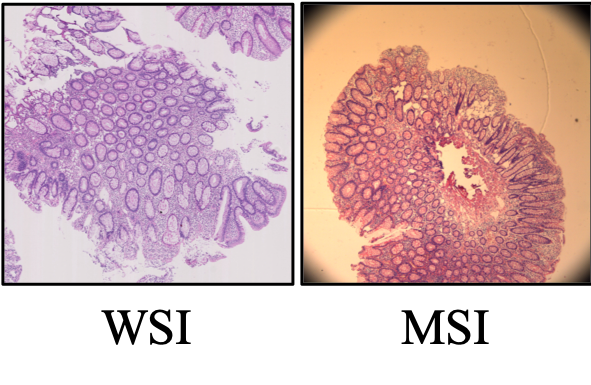}}
 \centerline{(b) background colors}
 \end{minipage} \hfill
  \begin{minipage}{0.3\linewidth}
 \centerline{\includegraphics[width=2.8cm]{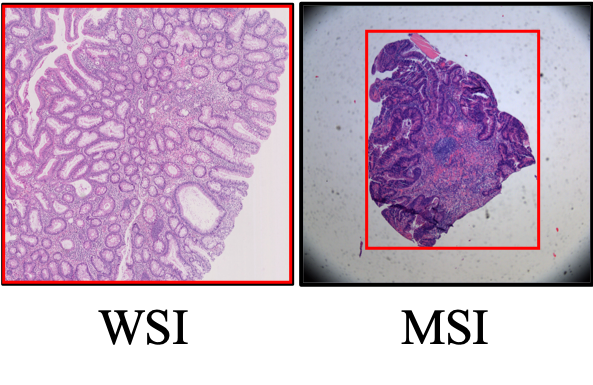}}
 \centerline{(c) field views}
 \end{minipage}
 \caption{Examples of domain discrepancies between whole-slide image (WSI) and microscopy image (MSI) on histopathological images, including scope shades, background colors and  field views. Here, the red boxes are for emphasis.}\label{variability}
 \vspace{-0.1in}
\end{figure}

\IEEEPARstart{D}{eep}  learning has achieved great success in various vision applications~\cite{jia2019deep,chen2020scripted,zhang2019whole,cao2019multi,yao2018deep}, such as image classification~\cite{chan2015pcanet} and resolution~\cite{lim2017enhanced}.
One widely accepted prerequisite of deep learning is rich annotated data~\cite{ref_partial1}.
In medical applications~\cite{dubey2015local,zhou2019high,khadidos2017weighted,zhang2020covid}, however, such rich supervision is often absent   due to prohibitive costs of data labeling~\cite{ref_survey2,ref_survey}, which impedes successful applications of deep learning.
Hence, there is a strong motivation to develop  {Unsupervised Domain Adaptation}  (UDA)~\cite{ref_domain_miccai,ref_transfer}  to improve diagnostic accuracy with limited annotated medical images.
Specifically, by leveraging a source domain with abundant labeled data,
UDA aims to learn a domain-invariant feature extractor,
which aligns the feature distribution of the target domain to that of the source.
As a result, the classifier trained with labeled source examples also  applies to unlabeled target  examples.  To achieve this, the key problem is to resolve the discrepancy between domains, derived from diverse characteristics of images from different domains (one  example can be found in Fig.~\ref{variability}).

\begin{figure}[t]
\centering
\includegraphics[width=8.5cm]{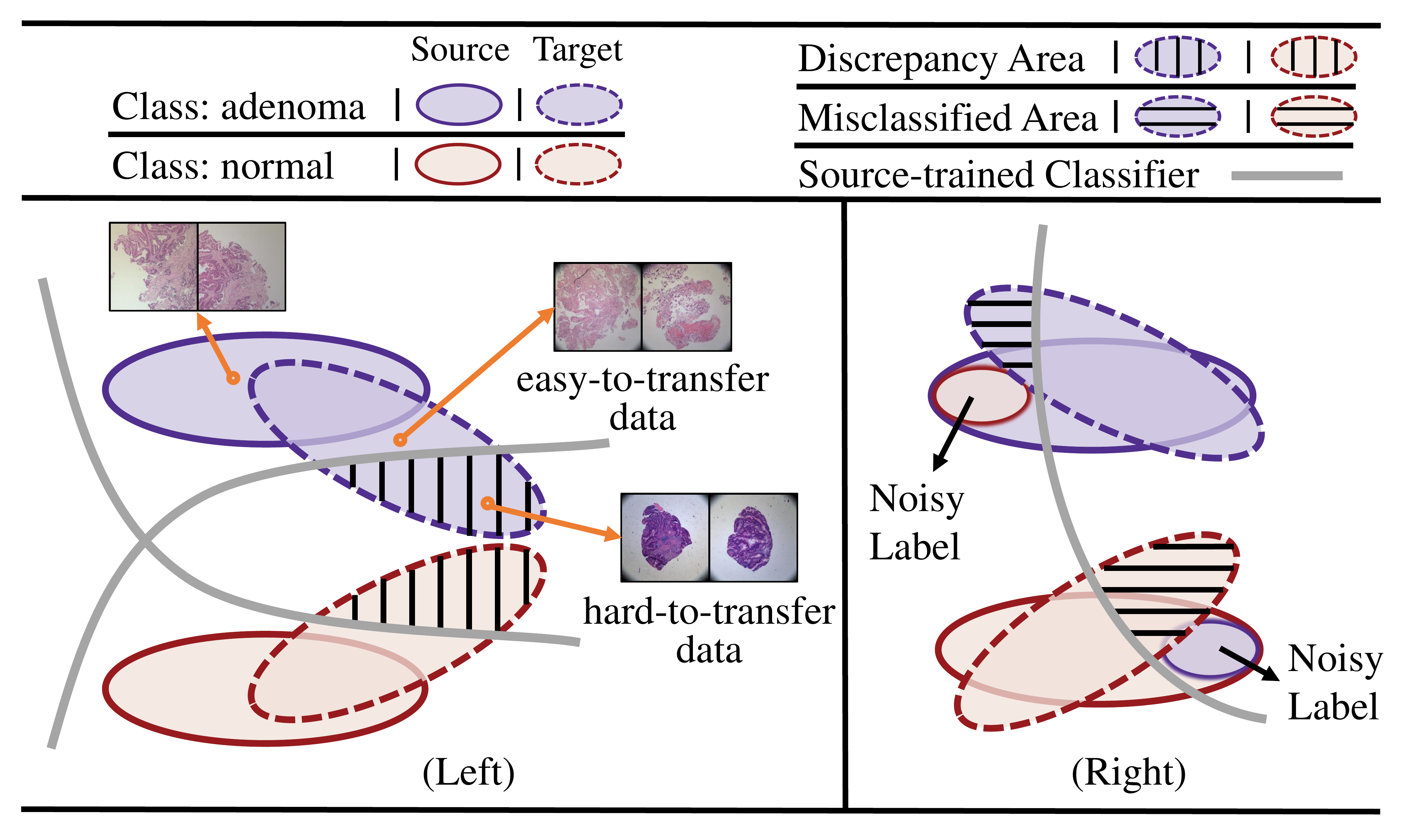} \vspace{-0.075in}
\caption{Illustrations of two challenges that UDA meets in medical image diagnosis, where discrepancy refers to prediction inconsistency of two classifiers.  Transferability difference (Left):  {hard-to-transfer  target data (dissimilar to the source) is potentially more distant to the source distribution, so they are more difficult to adapt than easy-to-transfer target data (similar to the source).} Label noise (Right): noisy labels on the source domain often result in a biased  {source-trained} classifier which inevitably performs poorly on the target domain.}
\label{motivation} \vspace{-0.1in}
\end{figure}

In addition to domain shift that all UDA methods resolve, medical image diagnosis poses two additional challenges.
\textbf{First}, there also exist significant discrepancies among images within the same domain.
The discrepancies, such as different appearance and scales of lesion regions,
mainly arises from inconsistent data preparation procedures, including tissue collection, sectioning and staining~\cite{zhang2019collaborative}.
As a result, different target images have different discrepancy levels to the source  images, and hence the difficulty of domain alignment  varies from sample to sample. That is, the target samples which bear a striking similarity with source samples are easier to align than the samples that are highly dissimilar (See the example in the left of Fig.~\ref{motivation}). If ignoring sample differences and  pursuing only global matching, those hard-to-transfer target images may not be well treated, leading  to inferior  overall  performance.
\textbf{Second}, a considerable percentage of medical annotations are unfortunately noisy, \yifan{caused by} diagnostic difficulties and subjective biases~\cite{ref_survey2}.
Directly applying the classifier built upon noisy source examples, as shown in the right of Fig.~\ref{motivation}, inevitably performs poorly on the target domain even though the target domain has been aligned well with the source. \textbf{Noteworthily}, the performance of UDA highly depends on both the domain alignment and the  classifier accuracy, so the mentioned two challenges are inescapable and are worth paying great attention to dealing with in the methodology of UDA for medical image diagnosis.

To address the two challenges that are usually ignored by existing UDA methods~\cite{ref_DANN,ref_DDC,ref_ADDA},
we innovatively propose a
Collaborative Unsupervised Domain Adaptation (CoUDA) algorithm.
By taking advantage of the collective intelligence of
two (or possibly more) peer networks~\cite{luo2019taking,ref_coteaching,ref_coteaching1},
CoUDA is able to distinguish between examples of different transferability (namely different levels of domain alignment difficulty).
Specifically, CoUDA assigns higher importance to the hard-to-transfer examples, which are collaboratively detected by peer networks  with greater prediction inconsistency (See prediction discrepancy in the left of Fig.~\ref{motivation}).
Meanwhile, we overcome label noise by a novel noise co-adaptation layer, shared between peer networks.
The layer aggregates different sets of noisy samples identified by all peer networks, and then denoises by adapting the predictions of these noisy samples to their (mislabeled) annotations. Last but not least, to further maximize the collective intelligence, we enforce the classifiers of peer networks to be diverse as large as possible.
Our main contributions are summarized as follows:

$\bullet$ We propose a novel Collaborative Unsupervised Domain Adaptation algorithm for medical image diagnosis.
Via the collective intelligence of two peer networks, the proposed method conducts transferability-aware domain adaptation and overcome label noise simultaneously.

$\bullet$ We theoretically analyze the generalization of the proposed method based on Rademacher complexity~\cite{ref_Rademacher_Bartlett}.

$\bullet$ We demonstrate the superiority and generalization of the proposed method through extensive experiments on both medical image and general image classification tasks.

 {\section{Related Work}} \label{2}
\textbf{Deep learning} has advanced many classification tasks for medical images~\cite{ref_survey2}. One prerequisite is the availability of labeled data, which is usually costly for medical diagnosis, since medical images can only be annotated by doctors with expertise~\cite{Hoi_2006}. Hence, there is a strong need to develop UDA to reduce labeling efforts in medical image tasks~\cite{ref_adaptation1,ref_adaptation2}.

\textbf{Unsupervised Domain Adaptation}. Existing UDA methods~\cite{ref_res,ref_cada,transferable_attention} for deep learning reduce the domain discrepancy either by adding adaptation layers to match high-order moments of distributions like  {Deep Domain Confusion} (DDC)~\cite{ref_DDC}, or by devising a domain discriminator to learn domain-invariant features in an adversarial manner, such as  {Domain Adversarial Neural Network} (DANN)~\cite{ref_DANN} and  {Maximum Classifier Discrepancy} (MCD)~\cite{ref_mcd}. Following the latter manner,  {Category-level Adversarial Network} (CLAN)~\cite{luo2019taking} further conducts category-level domain adaptation instead of only global alignment. This is similar to our idea, but we aim at a more fine-grained level, \ie sample-level difference of transferability. In addition, all these UDA methods ignore label noise and thus perform poorly in medical image diagnosis.

\textbf{Weakly-Supervised Learning against Noisy Labels}. Learning deep networks from noisy labels is an active research issue. Self-paced learning~\cite{ref_SPL} assumes samples with large losses are noisy and thus throws them away to keep clean. Following this method, MentorNet~\cite{ref_mentornet} learns a data-driven curriculum  to guide the training of base networks. Co-teaching~\cite{ref_coteaching,ref_coteaching1} learns two separate networks,
which guide the training of each other.
Another type of method tries to model the noise transition probability~\cite{Natarajan2013learning,ref_Patrini,ref_Sukhbaatar2014}.
In this way,  {Noise Adaptation Layer} (NAL)~\cite{ref_noiselayer} introduces an extra noise adaptation layer  to adapt network predictions to the noisy label, so that deep networks can well predict true labels.
Others methods try to adjust loss functions~\cite{dimensionality,bootstrap1,ref_bootstrap}.

\textbf{Unsupervised Domain Adaptation against Noisy Labels}. In this paper, we focus on UDA along with noisy labels in the source domain. To solve this,  {Transferable Curriculum Learning} (TCL)~\cite{ref_tcl} devises a transferable curriculum to improve domain adaptation via self-paced learning. Although it throws some noisy data away, it ignores class imbalance and also throws minority data with large losses away, leading to poor performance in practice. In addition, TCL also ignores different transferability of samples. In contrast, our method overcomes label noise without throwing important data  and thus is more general. Also, our method adaptively handle  hard-to-transfer data when adapting domains.

\textbf{Collaborative Learning}. Co-training~\cite{Blum1998} has been successfully applied in many learning paradigms, such as unsupervised learning~\cite{Kumar2011}, semi-supervised learning~\cite{Chen2011train,Zhou2005}, weakly-supervised learning~\cite{ref_coteaching} and multi-agent learning~\cite{multiagent}. In the methodology of UDA, one recent work~\cite{ref_coregularized} proposes a co-regularization scheme to improve the alignment of class conditional feature distributions. Moreover, tri-training~\cite{tri_training} can be regarded as an extension of co-training. Specifically, the work~\cite{tri_training1} proposes an asymmetric tri-training method for UDA, where they assign pseudo-labels to unlabeled samples and train neural networks using these pseudo-labels. However, all mentioned UDA methods ignore the issue of label noise and transferability differences, thus performing insufficiently in medical image diagnosis.\\

\begin{figure*}[t]
\centering
\includegraphics[width=17cm]{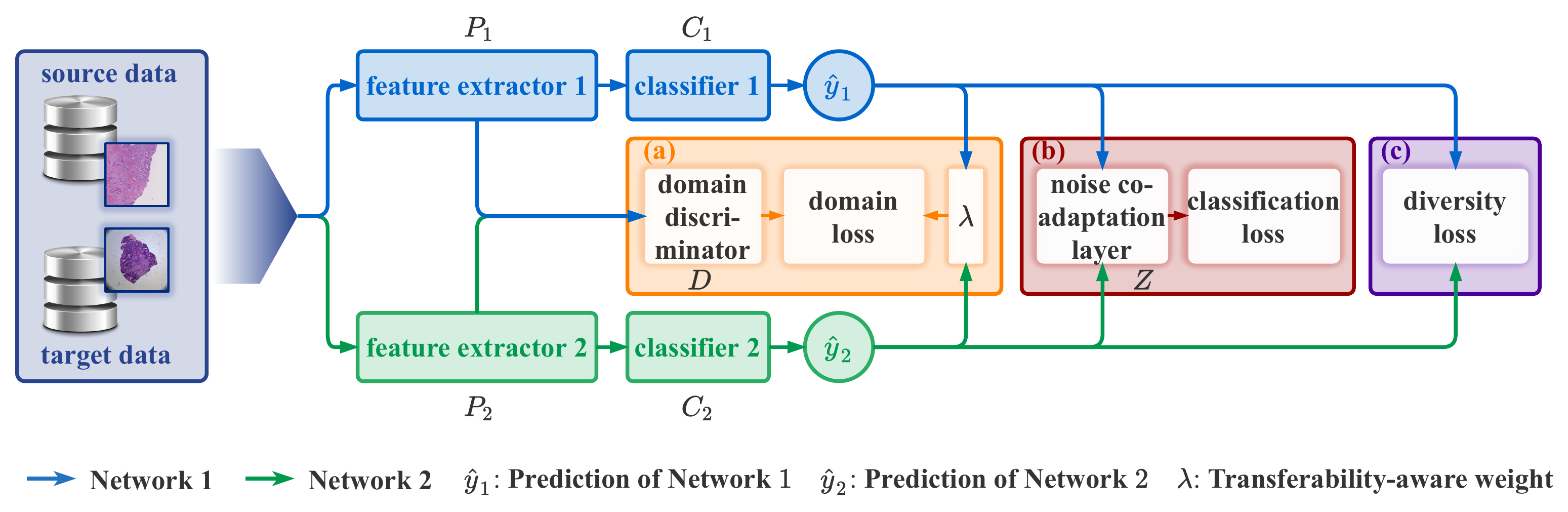}
\caption{The scheme of Collaborative Unsupervised Domain Adaptation (CoUDA). Two peer networks (in blue and green) have the same architecture but different parameters.  {In the training phase, both labeled source data (noisy) and unlabeled target data are fed into two peer networks for obtaining features and predictions. Based on the  predictions, there are three forward propagation branches: (a) we compute the  transferability-aware weight  $\lambda$ and use it to conduct transferability-aware domain adaptation; (b) we conquer label noise via a noise co-adaptation layer when conducting classification loss (only for labeled source data); (c) we maximize classifier diversity loss to make two peer networks keep diverse as large as possible. In the inference phase, the final prediction is the average prediction of two networks $\bar{y} \small{=}(\hat{y}_1 \small{+}\hat{y}_2)/2$.}} \label{framework}
\end{figure*}

\section{Method}

\textbf{Problem definition}. In this paper, we focus on the problem of  {Unsupervised Domain Adaptation} (UDA), where the model has access to only unlabeled samples
from the target domain. Formally, we denote $\mathcal{S}\small{=}\{x_i^s, y_i^s\}_{i=1}^{n_s}$ as the source domain of $n_s$ samples, where $x_i^s$ denotes the $i$-th sample and  $y_i^s \small{\in} \{0,1\}^K$ denotes its label, with $K$ being the number of classes.
In real-world medical diagnosis tasks, we may observe source samples with noisy labels, and we denote them as $\mathcal{S}'\small{=}\{x_i^s, z_i^s\}_{i=1}^{n_s}$.
In addition, the unlabeled target domain
$\mathcal{T}\small{=}\{x_j^t\}_{j=1}^{n_t}$ can be defined in a similar way.

 {The goal is to learn a well-performed deep model for the target  domain, using both noisy labeled source data and unlabeled target data. This task, however, is very difficult   due to (1) apparent discrepancies regarding domain distributions as well as data transferability, and (2) considerable label noise in the source domain. Most existing  UDA methods, however, resolve the domain discrepancy by assuming clean annotations or assuming samples are equally transferable, thus leading to limited performance in real-world medical applications. To address this task, we propose a novel Collaborative Unsupervised Domain Adaptation method, namely CoUDA.}

\subsection{Overall Collaborative Scheme of CoUDA}
There is an old saying that ``two heads are better than one", which actually highlights the importance of human cooperation in combating challenging tasks.
However, this is not true if the two heads are too weak or too  {similar}.
In other words, two heads should have strong abilities and also keep the diversity as large as possible, so that they can better deal with the task cooperatively.
 {Motivated by this,  we construct two separate peer networks\footnote{One can also construct more peer networks which may perform  better, but the learning cost will be a concern.  {Please refer to Appx. III in the supplementary for more details.}}  to overcome the domain discrepancy and label noise in a collaborative manner.
As shown in Fig.~\ref{framework}, the two networks have the same architecture (but different model parameters): a feature extractor $P_{\tau}$ for extracting domain-invariant features, and a classifier $C_{\tau}$ for prediction, where $\tau\small{\in\{1,2\}}$. In addition, the collaborative network consists of a domain discriminator $D$ for domain alignment and a noise co-adaptation layer $Z$ for handling label noise, where $D$ and  $Z$ are shared by two peer networks.}

 {To train the collaborative network, CoUDA conducts three main strategies:
(a)~\textbf{collaborative domain adaptation}: we cooperatively detect data transferability, and impose a new transferability-aware domain adversarial loss $\mathcal{L}^{d}$ to align the feature distribution of two domains, so that the domain discrepancy is minimized in an adversarial learning manner~\cite{zhang2019whole};
(b)~\textbf{collaborative noise adaptation}: we conquer label noise via the noise co-adaptation layer and train the whole network via a focal classification loss $\mathcal{L}^c$, which makes the model robust and imbalance-aware;
(c)~\textbf{classifier diversity maximization}: we maximize the diversity of two networks via a classifier diversity loss $\mathcal{L}^{div}$ so that peer networks can keep diverse as large as possible. In this way, CoUDA is able to adapt the source domain knowledge to the target
domain and diagnose medical images effectively.}

The overall training procedure is to solve the following minimax problem~\cite{ref_GAN}:
\begin{align}\label{objective}
 \min_{\theta_{p},\theta_{c},\theta_{z}}  \max_{\theta_{d}}~{\small{-}} \underbrace{\alpha\mathcal{L}^{d}(\theta_{p},\theta_{d})}_{\text{domain loss}} {\small{+}}\underbrace{\mathcal{L}^c(\theta_p,\theta_c,\theta_z)}_{\text{classification loss}}  {\small{-}} \underbrace{\eta \mathcal{L}^{div}(\theta_p,\theta_c)}_{\text{diversity loss}},
\end{align}
where $\theta_p\small{=}\{\theta_{p_1},\theta_{p_2}\}$ and $\theta_c\small{=}\{\theta_{c_1},\theta_{c_2}\}$ denote the parameters of the feature extractors $\{P_\tau\}_{\tau\small{=}1}^2$ and the classifiers $\{C_\tau\}_{\tau\small{=}1}^2$ in two peer networks, while $\theta_d$ and $\theta_z$ denote the parameters of the domain discriminator $D$ and the noise co-adaptation layer $Z$.
Here, $\alpha$ and $\eta$ denote trade-off parameters for the domain loss $\mathcal{L}^{d}$ and the diversity loss $\mathcal{L}^{div}$.

\subsection{Collaborative Domain Adaptation}\label{Sec_domain}

A key challenge for UDA is to resolve the domain discrepancy among images, while medical images usually have different potentials for domain alignment. Existing UDA methods often ignore such differences and treat all data equally. As a result, some hard-to-transfer images may not be well treated, leading to inferior performance.    To address this,
we propose to focus more on hard-to-transfer data.
The key question is how to detect the transferability of samples.

 {Previous work~\cite{ref_mcd} has shown that hard-to-transfer samples are often difficult to be correctly classified by two networks simultaneously.
Motivated by this and considering that two peer networks generate decision boundaries with different discriminability, we propose to detect data transferability collaboratively based on prediction inconsistency.
Formally,  for a sample $x_i$, we define the prediction inconsistency  based on the cosine distance, and model the transferability via a transferability-aware weight:}
\begin{align} \label{eq:weight}
  {\lambda(x_i)  = 2-\frac{\left<\hat{y}_{1,i}, \hat{y}_{2,i}\right>}{\|\hat{y}_{1,i}\|\|\hat{y}_{2,i}\|},}
\end{align}
 {where $\hat{y}_{1,i}$ and $\hat{y}_{2,i}$ denote the prediction probabilities of the sample $x_i$ by two classifiers, and $\left<a,b\right>\small{=}a^\top b$ denotes the inner product of two vectors $a$ and $b$. For simplicity, we represent $\lambda(x_i)$ by $\lambda_i$ in the following text}.

 {Based on this  transferability-aware weight, we conduct transferability-aware domain adaptation in an adversarial manner~\cite{ref_ADDA,ref_GAN}.}
 {Specifically, on the one hand, a domain discriminator $D$ is trained to adequately distinguish feature representations between two domains by minimizing the transferability-aware domain loss $\mathcal{L}^{d}$. On the other hand, the feature extractors of two peer networks are trained to confuse the discriminator by maximizing the domain loss.
In this way, the learned feature extractors are able to extract domain-invariant features that confuse the discriminator well.
Formally, based on the transferability-aware weight and least square distance~\cite{ref_LSGAN}, we define the domain loss as}:
 {
\begin{align}\label{eq:adversarial}
 \mathcal{L}^{d}(\theta_p,\theta_d) {\small{=}} \sum_{\tau=1}^2 {\Big{[}}   \frac{1}{n_s}\sum_{\mathcal{S}'}\lambda_i^s {d}_\tau(x_i^s)^2 {\small{+}} \frac{1}{n_t} \sum_{\mathcal{T}}\lambda_j^t (d_\tau(x_j^t){\small{-}}1)^2
 {\Big{]}},
\end{align}
where ${d}_{\tau}(x)\small{=}D(P_{\tau}(x))$ denotes the domain prediction by the domain discriminator $D$ and  feature extractor $P_\tau$.
Here, $\lambda_i^s$ and $\lambda_j^t$  denote the transferability-aware weights of the $i$-th source data and the $j$-th target data.}   Moreover, the label of the target domain is denoted as $1$ and that of the source is $0$.

\subsection{Collaborative Noise Adaptation} \label{NCN}

Label noise is common in medical image diagnosis.
If we directly match the predictions to noisy labels,  deep neural networks may even fit noisy samples due to strong fitting capabilities~\cite{rethinking}.
To make deep neural networks more robust, previous studies~\cite{ref_Patrini,ref_Sukhbaatar2014} assume that the noisy label $z$ is only conditioned on the true label $y$, \ie $p(z|y)$. After that, the prediction regarding the mislabeled class $m$ can be expressed as $p(z\small{=}m|x)\small{=}\sum_{k=1}^K p(z\small{=}m|y\small{=}k) p(y\small{=}k|x)$, where $p(z\small{=}m|y\small{=}k)$ means the probability that the true label $k$ changes to the noisy label $m$.  {Here, we use
a noise transition matrix to represent the transition probabilities regarding all classes.}
Following this, the work~\cite{ref_Sukhbaatar2014}  uses an additional noise adaptation layer to estimate the noise transition matrix  for deep networks.
However, the above assumption may not be    realistic in many cases, since a wrong annotation is more likely to occur in cases where features are misleading~\cite{ref_noiselayer}. Hence,
it is more reasonable to assume the noisy label $z$ is conditioned on both the true label $y$ and features $f$, \ie $p(z|y,f)$~\cite{ref_noiselayer}.

 {Based on this assumption, we propose a noise co-adaptation layer $Z$, shared by two peer networks. The motivation behind this collaborative design is that different networks have different discriminative abilities, and thus have different abilities to filter label noise out. That is, they can adjust the estimation error of noise transition probabilities, possibly caused by a single  network. Specifically, noise co-adaptation layer  estimates the probability} of true label $k$ changing to noisy label $m$ based on predictions and features using the softmax function:
\begin{align} \label{eq:softmax}
 p(\hat{z}{\small{=}}m|\hat{y}{\small{=}}k,f,\theta_z)=\frac{\exp{{\big{(}}{w^{\top}_{km}f+b_{km}{\big{)}}}}}{\sum_{l=1}^K \exp{{\big{(}}{w^{\top}_{kl}f+b_{kl}{\big{)}}}}},
\end{align}
where $\hat{z}$ denotes the prediction of the noisy label and $f$ represents the  features. Moreover, $w_{km}$ and $b_{km}$ denote the parameters of $Z$ regarding true label $k$ changing to noisy label $m$, and $\theta_z\small{=}\{w_{km},b_{km}|m,k\small{=}1,...,K\}$ denotes the parameter set of $Z$.
Based on Eq.~(\ref{eq:softmax}),  {for each peer network, we predict the noisy label  by:}
\begin{align}  \label{eq:predict_of_noisy_label}
  {p(\hat{z}{\small{=}}m|x,\theta_p,\theta_c\,\theta_z){\small{=}} \sum_k p(\hat{z}{\small{=}}m|\hat{y}{\small{=}}k,f,\theta_z)  p(\hat{y}{\small{=}}k|x,\theta_p,\theta_c).}
 \end{align}
In this way, we are able to train peer networks with noisy labels relying on any classification losses, \eg cross-entropy. In medical image diagnosis, since class imbalance is common, we adopt the focal loss~\cite{ref_Focal} as our classification loss:
\begin{align}\label{eq:classification}
 {\mathcal{L}^c(\theta_p,\theta_c,\theta_z) \small{=}\small{-} \frac{1}{n_s}\sum_{\mathcal{S'}} \sum_{\tau=1}^2    {z_{i}^s}^{\top} {\Big{(}}(1\small{-}\hat{z}_{\tau}(x_i^s) \big{)}^\gamma
  \log\big{(}\hat{z}_{\tau}(x_i^s)){\Big{)}}\,}
\end{align}
 {where $\hat{z}_{\tau}(x)\small{=}p(\hat{z}|x,\theta_{p_{\tau}},\theta_{c_{\tau}},\theta_z)$ denotes the prediction of the $i$-th source data and $z_{i}^s$ denotes its noisy label}, while $\gamma$ is a parameter to decide the degree to which the classification focuses on minority classes. Like~\cite{ref_Focal}, we set $\gamma\small{=}2$ in our method. By minimizing this loss, CoUDA focuses more on minority classes and thus handles class imbalance well.

\subsection{Classifier Diversity Maximization}\label{classifier_loss} %
It is worth mentioning that  the detection of data transferability highly depends on the classifier diversity. Also, keeping classifiers diverse prevents the noise co-adaptation layer reducing to a single noise adaptation layer in function~\cite{ref_noiselayer}. Hence, to better find hard-to-transfer samples out and handle label noise, we further maximize the classifier diversity. Specifically,  based on Jensen-Shannon (JS) divergence~\cite{ref_GAN}, we define the classifier diversity loss as:
\begin{align} \label{entropy_loss}
  {\mathcal{L}^{div}(\theta_p,\theta_c)  {\small{=}} \frac{1}{n_s{\small{+}}n_t}\sum_{\mathcal{S'}\cup\mathcal{T}} D_{KL}(\hat{y}_{1}\|\bar{y}) {\small{+}} D_{KL}(\hat{y}_{2}\|\bar{y}),}
\end{align}
where  $D_{KL}(y_1\|y_2)\small{=} \sum_i y_{1,i}\log(y_{1,i}) \small{-} y_{1,i} \log(y_{2,i})$ is the Kullback-Leibler divergence and $\bar{y} \small{=} \frac{\hat{y}_{1}\small{+}\hat{y}_{2}}{2}$.
Here, JS divergence is a symmetric distance metric, which measures the differences between two distributions well (See empirical superiority in Appx. III).
Moreover, since the final prediction of CoUDA is the average ensemble of two networks' predictions $\bar{y} \small{=} \frac{\hat{y}_{1}\small{+}\hat{y}_{2}}{2}$, maximizing the classifier diversity further improve ensemble performance~\cite{ref_ensemble}.

 {We summarize the training and inference details of CoUDA in Algorithm \ref{al:training} and Algorithm~\ref{al:inference}. Following~\cite{ref_DANN,ref_partial}, we implement the adversarial optimization with a gradient reversal layer (GRL)~\cite{ref_DANN}, which reverses the gradient of the domain loss when backpropagating to feature extractors.  In this way, the whole training process can be implemented with standard backpropagation in an end-to-end manner.}

\begin{algorithm}
    \small
    \caption{Training of CoUDA}\label{al:training}
    \begin{algorithmic}[1]
    \Require Noisy source data $\mathcal{S}'=\{x_i^s, z_i^s\}_{i=1}^{n_s}$; Target data $\mathcal{T}=\{x_j^t\}_{i=j}^{n_t}$; Training epoch $S$; Parameters $\alpha$ and $\eta$.
    \Ensure Feature extractors $\{P_1, P_2\}$;, Classifiers $\{C_1, C_2\}$; Noise co-adaptation layer $Z$; Domain discriminator $D$.
    \For{$s = 1 \to S$}
        \State Extract feature vectors $f_1, f_2$ based on $P_1, P_2$, respectively;
        \State Obtain the prediction $\hat{y}_1$ of Network 1 based on $C_1, f_1$;
        \State Obtain the prediction $\hat{y}_2$ of Network 2 based on $C_2, f_2$;
        \State Obtain transferability weight $\lambda$ based on $\hat{y}_1$ and $\hat{y}_2$;~~// Eq.~(\ref{eq:weight})
        \State Compute the domain loss $\mathcal{L}^{d}$ based on $f_1, f_2, D, \lambda$;~~// Eq.~(\ref{eq:adversarial})
        \State Compute the  diversity loss  $\mathcal{L}^{div}$ based on $\hat{y}_1$, $\hat{y}_2$.~~// Eq.~(\ref{entropy_loss})

        \If{Source data}
        \State Predict $\hat{z}_1$, $\hat{z}_2$ based on $\hat{y}_1, \hat{y}_2$, $Z$;~~// Eqs.~(\ref{eq:softmax}-\ref{eq:predict_of_noisy_label})
        \State Compute classification loss  $\mathcal{L}^c$ based on $\hat{z}_1$, $\hat{z}_2$.~~// Eq.~(\ref{eq:classification})
        \EndIf
        \State Compute the total loss;~~// Eq.~(\ref{objective})
        \State loss.backward(). ~~~//  standard backward propagation.
    \EndFor
     \end{algorithmic}
\end{algorithm}

\begin{algorithm}
\small
    \caption{Inference of CoUDA}\label{al:inference}
    \begin{algorithmic}[1]
    \Require Parameters of collaborative network: $P_1, P_2, C_1, C_2$;
    \Require Input data $x$.
    \State Compute the prediction $\hat{y}_1$ of Network 1  using $P_1, C_1, x$;
    \State Compute the prediction $\hat{y}_2$ of Network 2  using $P_2, C_2, x$;
    \State Obtain the ensemble prediction $\hat{y} = \frac{\hat{y}_1+\hat{y}_2}{2}$.
        \State Return: \hspace{0.5ex} $\hat{y}$.
    \end{algorithmic}
\end{algorithm}

\section{Theoretical Analysis}\label{Theorem}

This section analyzes the generalization error of the proposed CoUDA method based on Rademacher complexity~\cite{ref_Rademacher_Bartlett,Yu2018learning,Natarajan2013learning,Koltchinskii2002empirical}.
Before that, we give some necessary notations.

\textbf{Further Notations}: we denote  $\mathcal{\hat{S}}$, $\mathcal{\hat{S'}}$ and $\mathcal{\hat{T}}$ as the empirical distributions for $\mathcal{S}$, $\mathcal{S'}$ and $\mathcal{T}$, respectively.
In addition, we denote $l$ as the class labeling function, with $l_{\mathcal{S}}$ for the source domain, $l_{\mathcal{S'}}$ for the noisy source domain and $l_{\mathcal{T}}$ for the target domain.
For some hypothesis set $H$, let $h^*_{\mathcal{S}} \small{\in} {\rm argmin}_{h\in  H}\mathcal{L}_{\mathcal{S}}^{c}(h,l_{\mathcal{S}})$, $h^*_{\mathcal{S'}} \small{\in} {\rm argmin}_{h\in H}\mathcal{L}_{\mathcal{S'}}^{c}(h,l_{\mathcal{S'}})$, and $h^*_{\mathcal{T}} \small{\in} {\rm argmin}_{h\in H}\mathcal{L}_{\mathcal{T}}^{c}(h,l_{\mathcal{T}})$ be the optimal classifier regarding $\mathcal{S}$, $\mathcal{S'}$ and $\mathcal{T}$, respectively. Moreover, let $\hat{h}^*_{\mathcal{\hat{S}}} \small{\in} {\rm argmin}_{h\in  H}\mathcal{L}_{\mathcal{\hat{S}}}^{c}(h,l_{\mathcal{\hat{S}}})$ and $\hat{h}^*_{\mathcal{\hat{S}'}} \small{\in} {\rm argmin}_{h\in H}\mathcal{L}_{\mathcal{\hat{S}'}}^{c}(h,l_{\mathcal{\hat{S}'}})$ be the optimal classifiers regarding two empirical source distributions.
Also, we denote the \ming{expected risk over}  $\mathcal{X}\small{=}\{X,Y\}$ as $R_{\mathcal{X}}(h)\small{=}\mathbb{E}_{\mathcal{X}}\mathcal{L}_{\mathcal{X}}^{c}(h(X),Y)$, and the empirical risk by $\hat{R}_{\mathcal{X}}(h)\small{=}\frac{1}{n}\sum_{i=1}^n\mathcal{L}_{\mathcal{X}}^{c}(h(x_i),y_i)$.

 {Following~\cite{Chazelle2001the,Mansour2009domain}, we define the discrepancy distance between the source distribution $\mathcal{S}$ and the target distribution $\mathcal{T}$ as:
$ {\rm disc}_{\mathcal{L}}(\mathcal{S},\mathcal{T})\small{=}\max_{h_1,h_2\in H}\big{|} \mathcal{L}_{\mathcal{S}}(h_1,h_2)\small{-}\mathcal{L}_{\mathcal{T}}(h_1,h_2)\big{|}$, where $H$ means a set of hypothesis, and $\mathcal{L}$ denotes some loss function.
We then have the following result on the generalization error.}
\begin{theorem}\label{theorem}
 {Let $U$ be a hypothesis set for the domain loss $\mathcal{L}^{d}$ and $H$ be a hypothesis set for the classification loss $\mathcal{L}^{c}\in [0,M]$ in terms of $K$ classes. Assume that the loss function $\mathcal{L}$ is symmetric and obeys the triangle inequality. Suppose the noise transition matrix $Q$ is invertible and known.  For any $\delta>0$ and any hypothesis $h\in H$, with probability at least $1-(2\small{+}K)\delta$ over $m$ samples drawn from $\mathcal{S}$ and $n$ samples drawn from $\mathcal{T}$,  the following holds:}
\begin{align}
  &\mathcal{L}_{\mathcal{T}}^{c}(\hat{h}^*_{\mathcal{\hat{S}'}},l_{\mathcal{T}}) -  \mathcal{L}_{\mathcal{T}}^{c}(h^*_{\mathcal{T}},l_{\mathcal{T}}) \nonumber \\
  & \leq {\rm disc}_{\mathcal{L}^{d}}(\mathcal{\hat{S}},\mathcal{\hat{T}})  +\mathcal{L}_{\mathcal{S}}^{c}(h^*_{\mathcal{S}},h^*_{\mathcal{T}})
  + 4K\mathfrak{\hat{R}}_{\mathcal{S}}(H) \nonumber \\
   &+8 \mathfrak{\hat{R}}_{\mathcal{S}}(U)+8\mathfrak{\hat{R}}_{\mathcal{T}}(U) \small{+}9\Bigg{(}(1\small{+}\frac{2}{3}M)\sqrt{\frac{\log \frac{2}{\delta}}{2m}} \small{+}\sqrt{\frac{\log \frac{2}{\delta}}{2n}}\Bigg{)}.\nonumber
\end{align}
\end{theorem}
See Appx. I for the proof and the definition of the empirical Rademacher complexity $\mathfrak{\hat{R}}(\cdot)$. This theorem shows the generalization error of CoUDA.
 {Overall, if we  minimize the domain discrepancy ${\rm disc}_{\mathcal{L}^{d}}(\mathcal{\hat{S}},\mathcal{\hat{T}})$ well, the learned classifier on the source domain can match the optimal classifier on the target well. This verifies the reasonability and significance to reduce domain discrepancy ${\rm disc}_{\mathcal{L}^{d}}(\mathcal{\hat{S}},\mathcal{\hat{T}})$ via transferability-aware domain adaptation. Moreover, $\mathcal{L}_{\mathcal{S}}^{c}(h^*_{\mathcal{S}},h^*_{\mathcal{T}})$ denotes the average  loss  between  the  best  intra-class  hypothesis, and is usually assumed to be small~\cite{Mansour2009domain}. This is because if there is not any hypothesis that performs well on both domains, domain adaptation cannot be conducted.}

\section{Experimental Results}\label{experiment}

\begin{table*}[t]
\vspace{-0.1in}
	\caption{ {Comparisons on colon cancer  diagnosis in terms of four metrics.}}
	\label{performance}
	\vspace{-0.1in}
 \begin{center}
 \begin{threeparttable}
    \resizebox{0.8\textwidth}{!}{
 	\begin{tabular}{lcccc|cccc}\toprule
        \multirow{2}{*}{Methods}&\multicolumn{4}{c|}{Colon-A}&\multicolumn{4}{c}{Colon-B}\cr\cmidrule{2-5}\cmidrule{6-9}
        & Acc (\%) & MP & MR  & Macro F1  & Acc (\%)& MP &  MR  & Macro F1 \cr
        \midrule
         MobileNet~\cite{ref_Mobilenet}  &68.79	 	& 78.62	& 61.67	& 64.46         &44.84	 	& 61.08	& 40.77	& 33.72\\
         MentorNet~\cite{ref_mentornet}  &66.28	 	& 79.24 	& 55.65	& 56.25       &41.25	 	& 65.60& 42.84	& 37.06\\
         Co-teaching~\cite{ref_coteaching}  & 71.52 &79.74	& 64.88	& 67.93             &48.16	 	& 74.73	& 46.74	& 43.42\\	
         NAL~\cite{ref_noiselayer}&68.92 	& 74.83	& 63.49	& 64.65              &50.23	 	& 54.78	& 43.32	& 40.71 \\
         \midrule
         DDC~\cite{ref_DDC} & 79.74 & 78.75	& 79.01 		& 78.76	           &65.21	 	&65.76	& 68.82	&65.64\\
         DANN~\cite{ref_DANN} & 80.01 	& 79.25	& 78.85 	& 78.75              &67.23	 	& 67.40 & 67.63	& 66.98\\
         MCD~\cite{ref_mcd} &81.77 	& 81.04	& 80.45	& 80.19                  &68.20	 	& 67.46	& 66.92	& 66.86 \\
         CLAN~\cite{luo2019taking} &82.47 	& 80.75	& 81.69	& 81.04                  &72.96	 	& 75.93	& 73.88	& 73.35 \\
         TCL~\cite{ref_tcl} &81.26 	& 80.36	& 80.97	& 80.27                  &69.24	 	& 69.93	& 69.53	&  68.75 \\
          \midrule
         CoUDA({\scriptsize{Source-only}}) &72.51 	& 77.52	&66.65	&69.24                  &53.05	 	& 65.35	& 50.78	& 47.89 \\
          CoUDA  & \textbf{87.75} 	& \textbf{87.62}	& \textbf{86.85} 	& \textbf{87.22}     &    \textbf{78.35} 	& \textbf{78.37}	& \textbf{79.48} 	& \textbf{78.63}     \\
        \bottomrule

	\end{tabular}}
	  \vspace{-0.1in}
	 \end{threeparttable}
	 \end{center}
\end{table*}

\subsection{Experimental Settings}
\textbf{Baselines}.
We compare \textbf{CoUDA} with several advanced methods, including:
(a) \textbf{MobileNet} V2~\cite{ref_Mobilenet} is the network backbone.
(b) \textbf{MentorNet}~\cite{ref_mentornet}, \textbf{Co-Teaching}~\cite{ref_coteaching} and \textbf{NAL}~\cite{ref_noiselayer} are classic weakly-supervised learning methods for training deep networks.
(c) \textbf{DDC}~\cite{ref_DDC}, \textbf{DANN}~\cite{ref_DANN}, \textbf{MCD}~\cite{ref_mcd} and \textbf{CLAN}~\cite{luo2019taking} are classic unsupervised domain adaptation methods.
(d) \textbf{TCL}~\cite{ref_tcl} is the state-of-the-art unsupervised domain adaptation method that considers  label noise.
(e) \textbf{CoUDA (Source-only)} is a variant of CoUDA with only classification loss on source samples (without domain adaptation).
For fairness, all methods employ the same network backbone yet with different optimization procedures.

\textbf{Implementation Details}.
We implement CoUDA with Tensorflow\footnote{The source code is available: https://github.com/Vanint/CoUDA.}.
For medical images, we train the model from scratch, while for general images we use MobileNet pre-trained on the ImageNet~\cite{ref_imagenet} as the backbone.
In the training process, we use Adam optimizer with the batch size 16 on a single GPU. The learning rate is set to $10^{-5}$ for medical tasks and $10^{-3}$ for tasks with general images. Moreover, we set the trade-off parameters $\alpha \small{=} 0.1$ and $\eta \small{=} 0.01$.  {The overall training step is $10^5$. More implementation details  about the network architecture and noise co-adaptation layer (\eg initialization) can be found Appx. V.B.1.}

\textbf{Metrics}.
In this paper, we use Accuracy (Acc), Macro Precision (MP), Macro Recall (MR) and Macro F1-measure as metrics. Their implementations can be found in~\cite{metric}.

\subsection{ {Evaluation on Colon Cancer Diagnosis}}

\textbf{Dataset}.
The colon dataset is formed by H$\&$E stained  histopathology slides, which are diagnosed as 4 types of colon polyps (normal, adenoma, adenocarcinoma, and mucinous adenocarcinoma). This dataset is provided by Tencent AI Lab and data statistics are shown in Table~\ref{dataset}. Here, the whole-slide image (WSI) is regarded as the source data, while the microscopy image (MSI) is viewed as the unlabeled target data. Specifically, WSIs of each slide is acquired in 40\(\times\) magnification scale (229 nm/pixel) by the Hamamatsu NanoZoomer 2.0-RS scanner, where ROIs (\ie region of interest) corresponding to the categories are then annotated by experts on WSI scans with our in-house tool. MSIs of 30 slides are acquired with microscope in 10\(\times\) magnification scale (field of vision: {\(2.73\small{\times}2.73\)} cm\textsuperscript{2}, matrix size: {\(2048\small{\times}2048\)}). Specifically, we focus on 10\(\times\) magnification scale, since it is the preferred scale for pathologist's diagnosis.
For consistency, WSIs are down-sampled to the same resolution as MSIs. Following that, a sliding window crops MSIs and annotated WSI regions into patches of size 512, where the label of each 
patch is defined by the annotation in its center. WSI and MSI patches acquired from 15 slides are taken as the test set, and the rest serves as the training set.

From Table~\ref{dataset}, we find class imbalance is severe.
Moreover,  {according to the remark of doctor annotators, there exist a certain number of noisy labels in WSIs due to coarse annotation, while the labels of MSIs in the test set are clean based on careful annotation. Considering both class imbalance and label noise, this problem is very tough. For adequate evaluations, we formulate two multi-class classification tasks for domain adaptation, where the \textbf{Colon-A} task consists of three classes (normal, adenoma and adenocarcinoma), while \textbf{Colon-B} consists of all four classes. Generally, Colon-B is more difficult than Colon-A, since it suffers from more severe class imbalance and label noise due to the additional class.}

\begin{table}[t]
	\caption{ {Statistics of colon image dataset, where WSI serves as the source domain, and MSI serves as the target domain. In addition, ``ade." denotes the adenocarcinoma class.}}\label{dataset}
     \begin{footnotesize}
    \begin{center}
          \begin{threeparttable}
       \renewcommand\arraystretch{1}
        \renewcommand{\tabcolsep}{3.5pt}
        \resizebox{0.48\textwidth}{!}{
	\begin{tabular}{ccccccc}\toprule
        \multirow{2}{*}{Set} &\multirow{2}{*}{Domain} &\multicolumn{4}{c}{Categories}&\multirow{2}{*}{total} \\    \cmidrule{3-6}
          & & normal& adenoma& ade. & mucinous ade. & \cr

        \midrule
        \multirow{2}{*}{Training} &
        WSI          &36,094 	&3,626 		& 3,081 		& 1,741  & 44,542  \\
        &MSI    &2,696  	& 1,042	 	& 1,091 		&2,534 	& 7,363	\\
        \midrule
        Test &
        MSI   	& 1,110 	& 487 & 713 	& 674  & 2,984	\\
        \bottomrule
	\end{tabular}}
	  \end{threeparttable}
    \end{center}
    \end{footnotesize} \vspace{-0.1in}
\end{table}

\begin{table*}[t]
\centering
\caption{ {Ablation studies of CoUDA on Colon-A. $\mathcal{L}^{ce}$ denotes cross entropy classification loss; TW denotes transferability-aware weight; NCL denotes the noise co-adaptation layer.}}
\label{Ablation}
  \renewcommand{\tabcolsep}{9.0pt}
 \scalebox{1}{
\label{module-eval}
\newcommand{\tabincell}[2]{\begin{tabular}{@{}#1@{}}#2\end{tabular}}
\begin{tabular}{ccccc|cc|cccc}
\toprule
 \tabincell{c}{Backbone} & ~$\mathcal{L}^{ce}$~ & $\mathcal{L}^{c}$ & $\mathcal{L}^{d}$ & $\mathcal{L}^{div}$ & TW & NCL & Acc ($\%$) & MP & MR & Macro F1\\
\midrule
$ \surd$ & $\surd$ & & & &  &  &  69.57 & 79.21 & 58.53	& 59.24  \\
  $\surd$ &  & $\surd$ & & & $\surd$ & $\surd$ &  72.51 & 77.52	& 66.65 & 69.24 \\
 $\surd$ & & $\surd$ & $\surd$ & & $\surd$ & $\surd$   &85.54 	& 84.49	&85.22	& 84.81  \\
$\surd$  & & $\surd$ & $\surd$ & $\surd$ & $\surd$ &   &82.94 	& 81.37	&82.70	& 81.64   \\
$\surd$  & & $\surd$ & $\surd$ & $\surd$ &  & $\surd$   &85.45 	& 84.46	&84.94	& 84.68  \\\midrule
  $\surd$& & $\surd$ & $\surd$ & $\surd$ & $\surd$ & $\surd$   & \textbf{87.75} 	& \textbf{87.62}	& \textbf{86.85} 	& \textbf{87.22}    \\

\bottomrule
\end{tabular}
}
\end{table*}

\begin{figure*}[ht]
\vspace{0.1in}
 \begin{minipage}{1\linewidth}
 \centerline{\includegraphics[width=12cm]{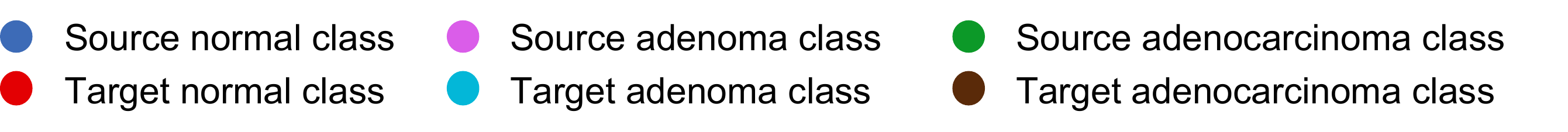}}
 \end{minipage}
 \vfill \vspace{0.05in}
 \begin{minipage}{0.32\linewidth}
 \centerline{\includegraphics[width=4.3cm]{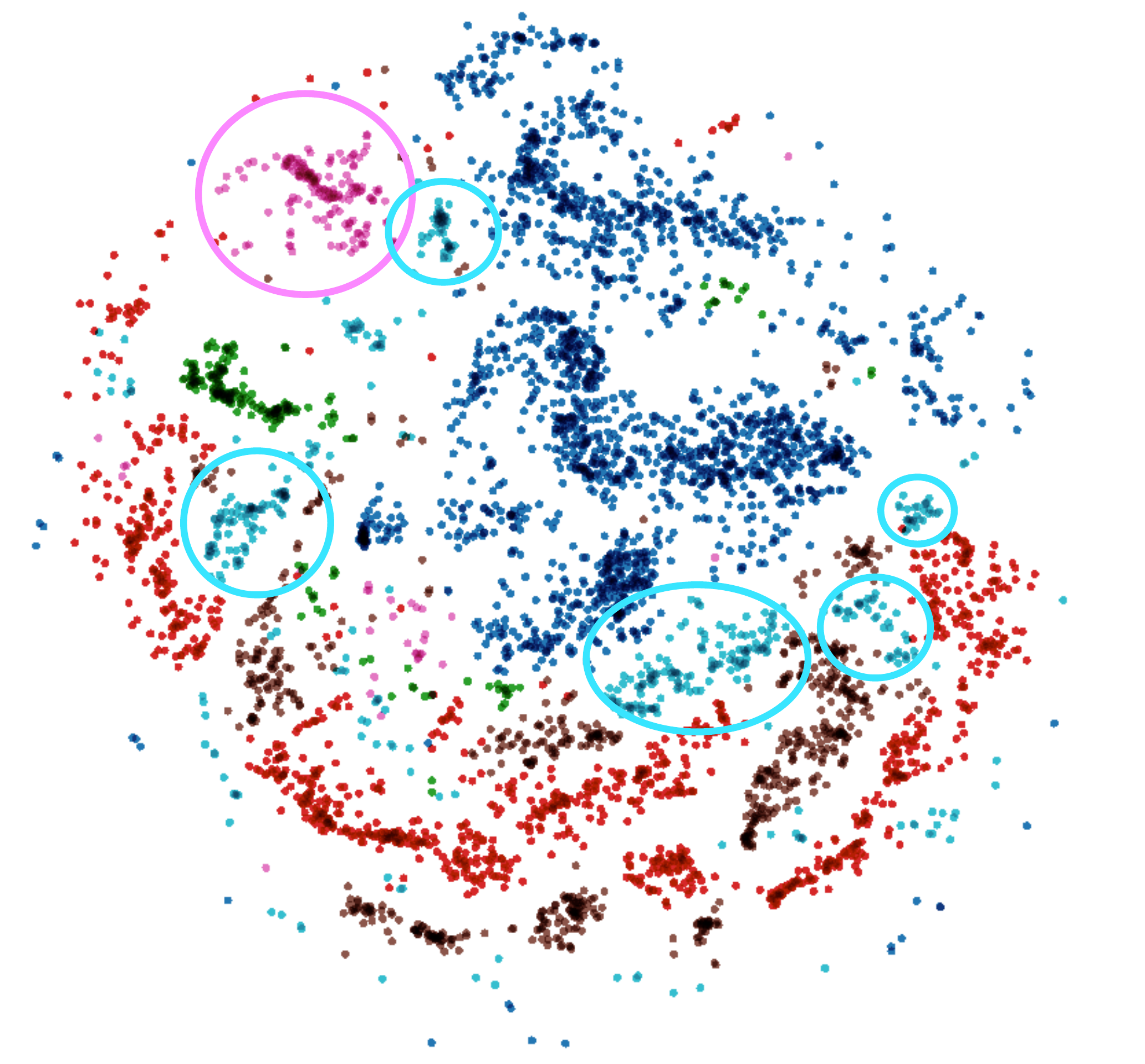}}
 \centerline{(a) MobileNet}
 \end{minipage}
 \hfill
 \begin{minipage}{0.32\linewidth}
 \centerline{\includegraphics[width=4.3cm]{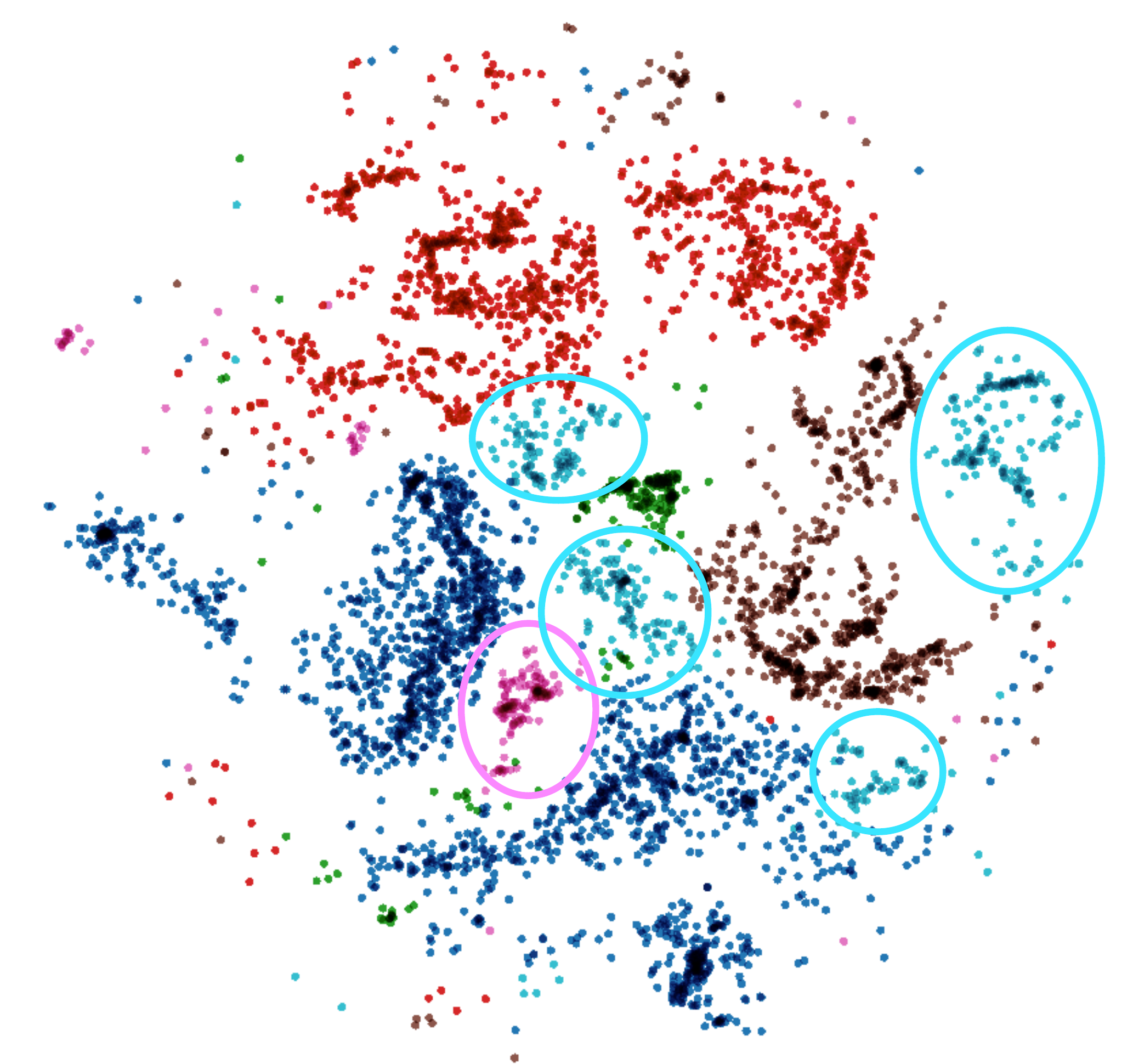}}
 \centerline{(b) DANN}
  \end{minipage}
  \hfill
 \begin{minipage}{0.32\linewidth}
 \centerline{\includegraphics[width=4.3cm]{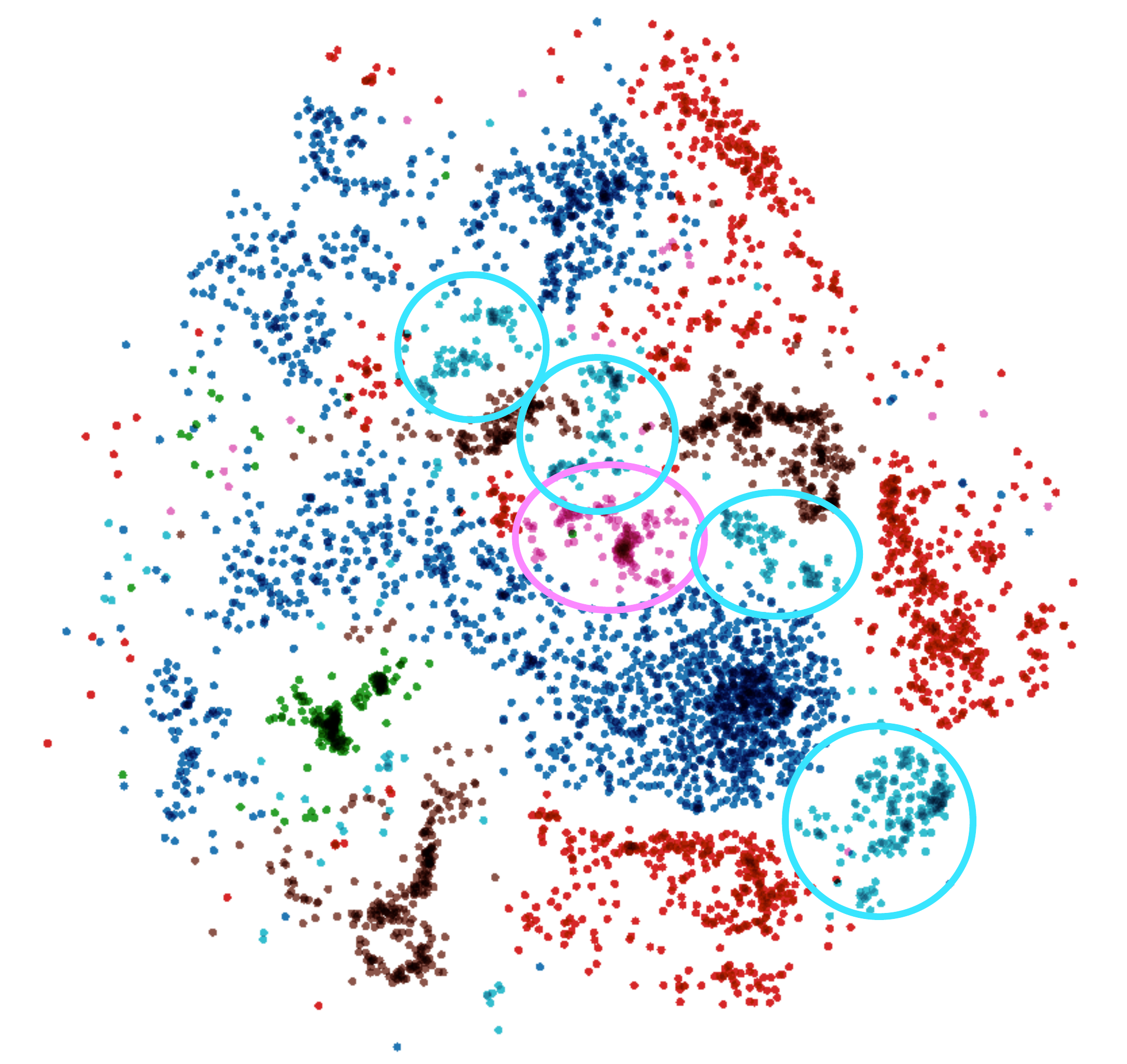}}
 \centerline{(c) TCL}
 \end{minipage}
 \vfill
 \begin{minipage}{0.32\linewidth}
 \centerline{\includegraphics[width=4.cm]{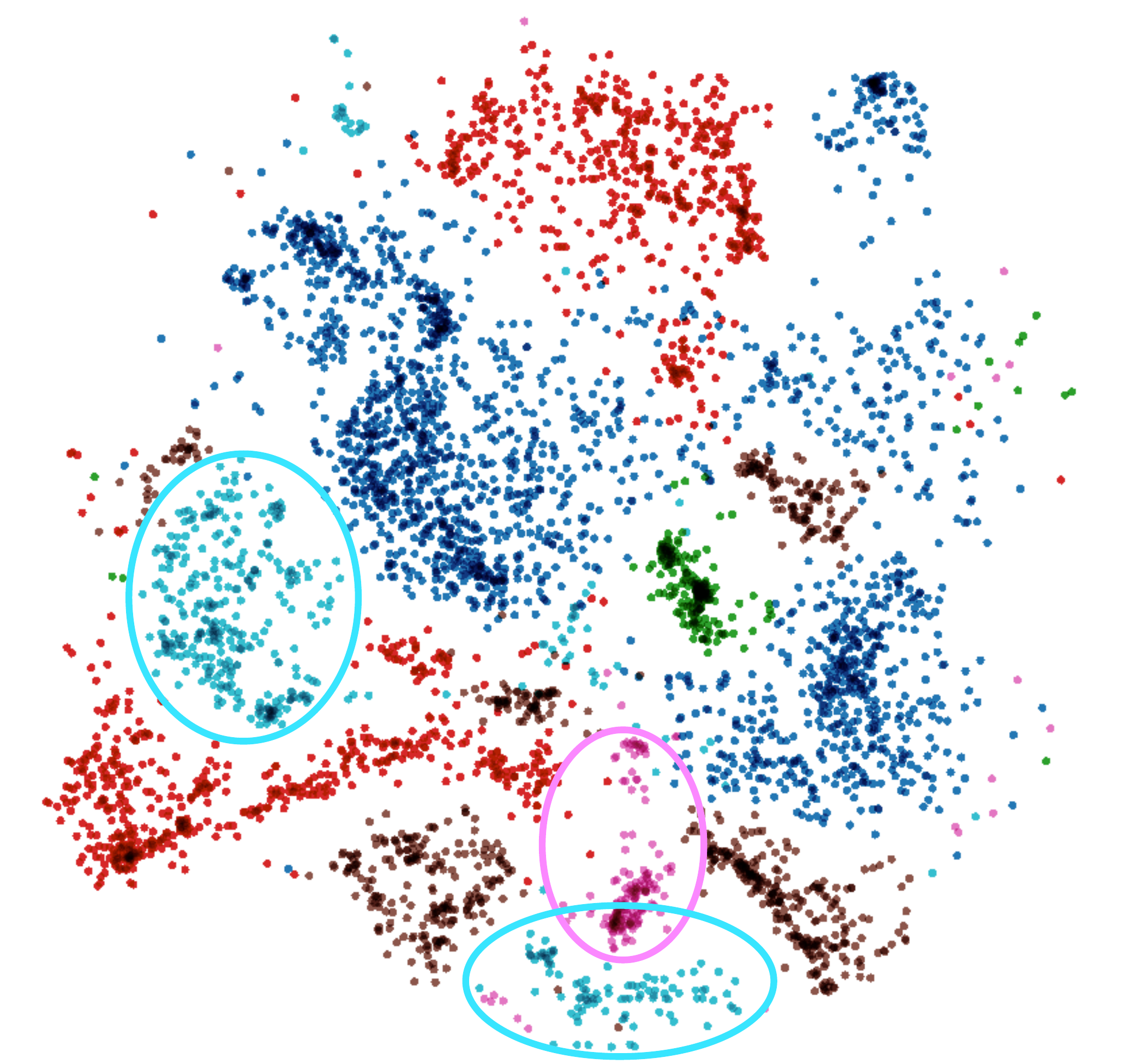}}
 \centerline{(d) MCD}
 \end{minipage}
 \hfill
  \begin{minipage}{0.32\linewidth}
 \centerline{\includegraphics[width=4.3cm]{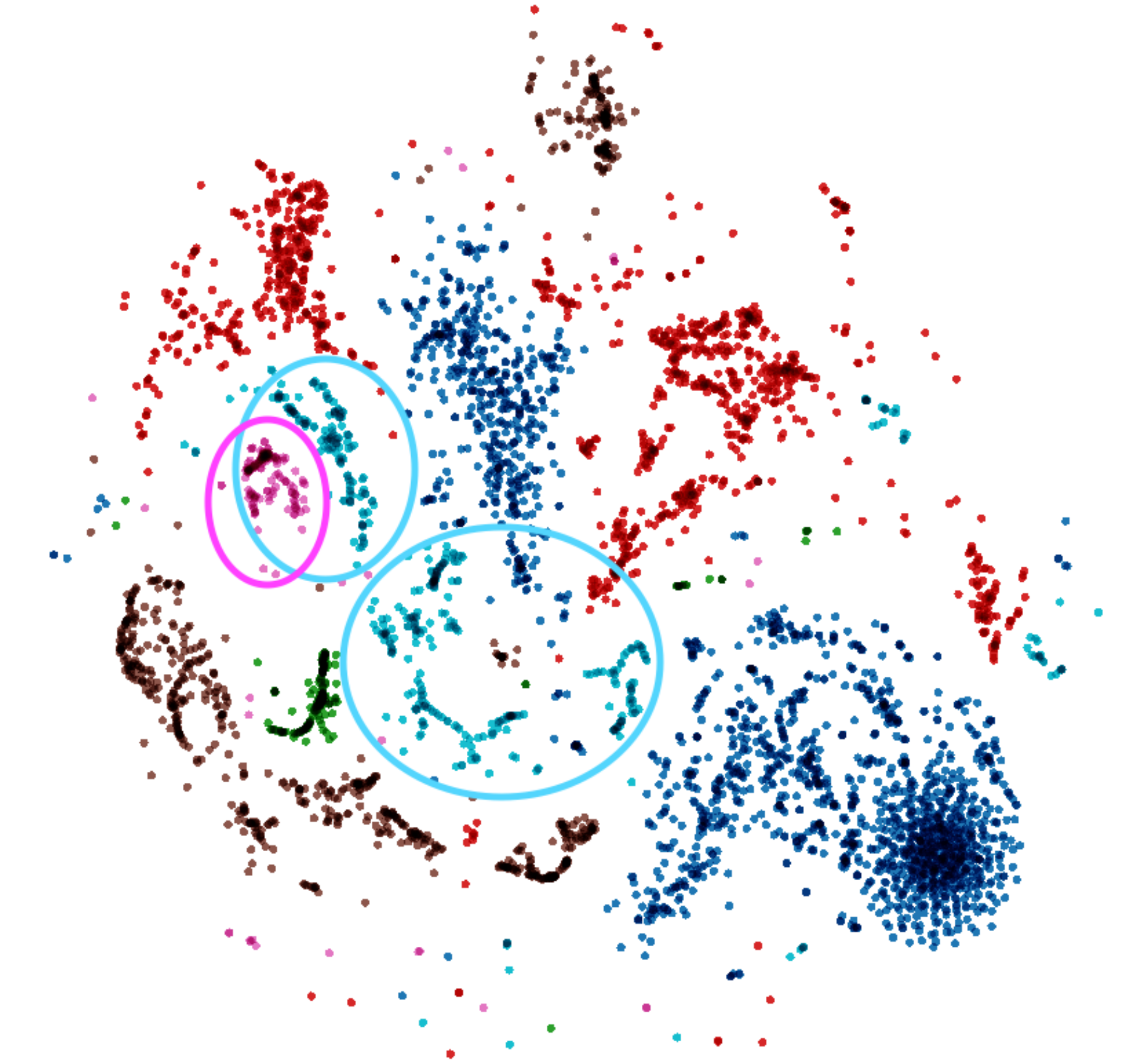}}
 \centerline{(e) CLAN}
 \end{minipage}
 \hfill
 \begin{minipage}{0.32\linewidth}
 \centerline{\includegraphics[width=4.3cm]{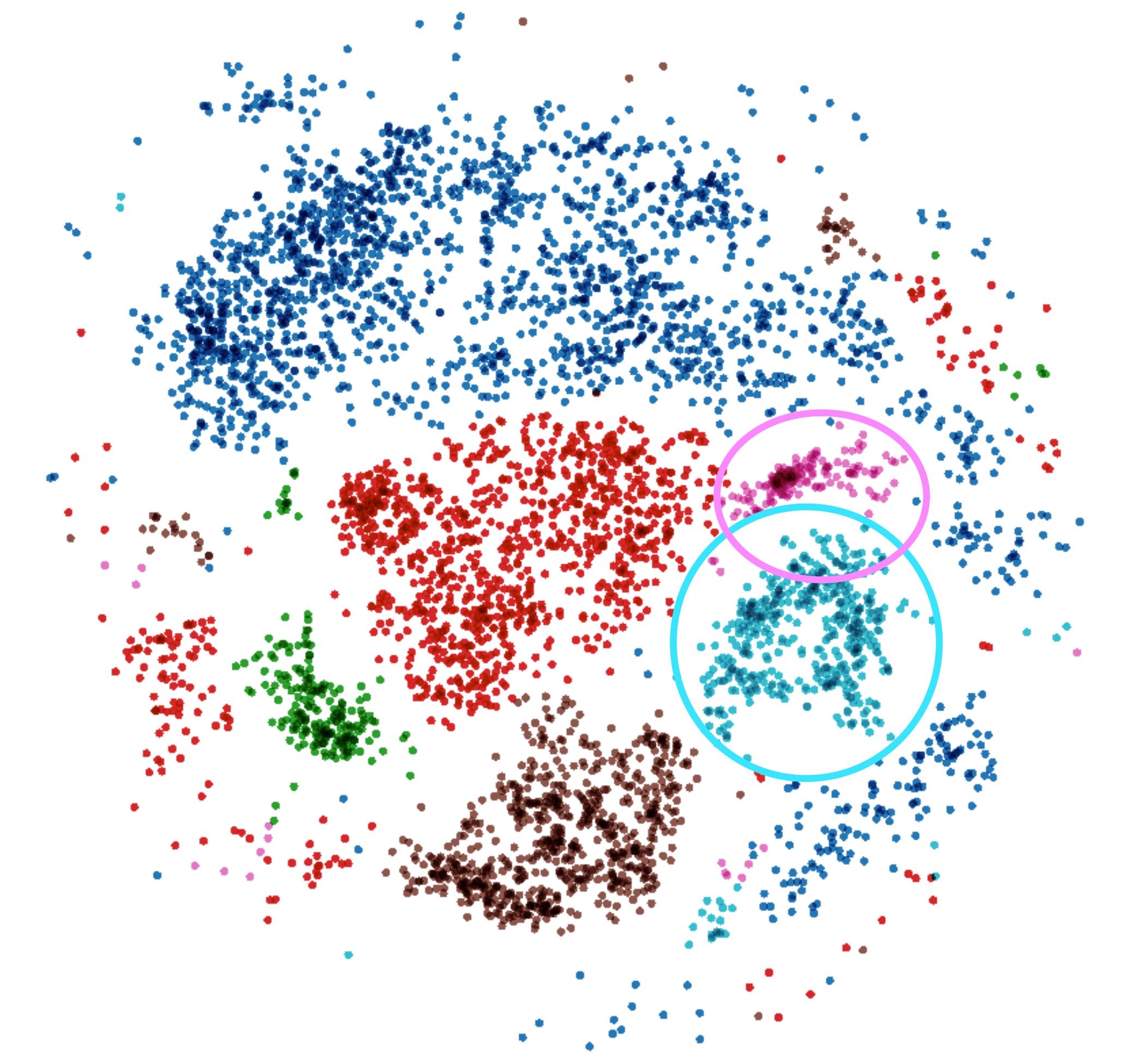}}
 \centerline{(f) CoUDA}
 \end{minipage}
 \caption{\yifan{t-SNE plots of learned features on Colon-A in terms of classes and domains, where the clusters of adenoma are circled. The closer the same class samples across domains are, the more effective the domain adaptation is (better viewed in color).}
 }
 \label{tsne}
\end{figure*}

\textbf{Results}.
 {We report the result of all methods   in Table~\ref{performance}.
The encouraging result  suggests that CoUDA is able to conduct effective domain adaptation from the source domain (with label noise) to the unlabeled target domain.
Since there is an urgent need for robust deep models for medical image diagnosis, this result highlights the importance of CoUDA in real-world medical applications with label limitations.}

 {From the empirical result, we further make the following observations. (1) The issue of class imbalance in medical image diagnosis makes the learning  of deep models from noisy annotations more difficult. For example, MentorNet proposes to discard noisy data  with large losses. However, it also discards the minority data that often suffers from large losses~\cite{zhang2018online}, leading to inferior performance than MobileNet due to class imbalance. In contrast, NAL handles the label noise and diagnoses the class-imbalanced cancer images better, since the noise adaptation layer does not throw the minority  data away.
(2) The collaborative scheme helps to handle label noise better. For example, Co-teaching performs better than MenterNet, while CoUDA (source-only) with noise co-adaptation layer outperforms NAL. (3) Reducing the domain discrepancy by learning domain-invariant features can well adapt the source domain knowledge to the target domain  (DDC, DANN, MCD and CLAN). This also demonstrates the importance of domain adaptation in medical image diagnosis.
(4) Simultaneously resolving the domain discrepancy and label noise makes great performance contributions (CoUDA and TCL). However, TCL follows the technique of MentorNet and ignores the issue of class imbalance.}

\textbf{Ablation Studies}.
We then conduct ablation studies on Colon-A. As shown in Table~\ref{Ablation},  {promising results verify the effectiveness of  all components in our methods, where each of them makes empirical contributions. In addition, according to the result, the transferability-aware domain loss for  domain adaptation  and the noise co-adaptation layer  for noise adaptation are relatively more important. Such a result is reasonable, since the issues of domain discrepancy and label noise are the main challenges in this task.}

\textbf{Feature Visualization}.
We \normaljudy{visualize} t-SNE embeddings~\cite{ref_tsne} of \normaljudy{learned features after} the global average pooling. In our collaborative network, we only plot the features learned from one peer network.  {By taking the adenoma class as an example, we make several interesting observations. (1) There are apparent domain discrepancies on Colon-A, where the feature distributions of two domains are quite different (See Fig.~\ref{tsne}(a)).
(2) Ignoring the data difference regarding transferability may make  domain alignment difficult. For example, DANN, TCL, MCD and CLAN
fail to align domains very well (See Figs.~\ref{tsne}(b-e)).
(3) The cooperation scheme is helpful for domain alignment. For instance, CoUDA aligns  domain features  well by conducting collaborative domain adaptation. These visualized results further explain the superiority of CoUDA in domain knowledge adaptation.
}

\textbf{Parameter Sensitivities}.
As mentioned in Section V.A, we set the trade-off parameter $\alpha \small{=} 0.1$ and $\eta \small{=} 0.01$ in all experiments, where $\alpha$ adjusts the domain loss and $\eta$ controls the diversity loss.
In this section, we further evaluate the sensitivities of the two parameters  {on Colon-A. In each experiment, we only evaluate one parameter, while fixing all other parameters. We report the results in Table~\ref{sensitivities}, which shows that  the proposed method is insensitive to both parameters  in the range of $\{10^{-3},10^{-2},10^{-1},1\}$. Moreover, the proposed method achieves the best on Colon-A when setting $\alpha\small{=}0.1$ and $\eta\small{=}0.01$. In other real-world applications, it would be better for users to select suitable values based on the task in hand.}

\begin{table}[h]
	\caption{ {Parameter sensitivities analysis on Colon-A.}}
	\label{sensitivities}
 \begin{center}
      \begin{threeparttable}
       \renewcommand\arraystretch{1}
        \renewcommand{\tabcolsep}{3.5pt}
        \resizebox{0.48\textwidth}{!}{
 	\begin{tabular}{lcccc|cccc}\toprule
      \multirow{2}{*}{Parameters}&\multicolumn{4}{c|}{$\alpha$}&\multicolumn{4}{c}{$\eta$}\cr\cmidrule{2-5}\cmidrule{6-9}
        & 0.001  &0.01 & 0.1  &1   & 0.001  &0.01 & 0.1  &1   \cr
        \midrule
         Acc (\%)  &87.23	& 87.53	&\textbf{87.75}	& 86.36        &87.62 	& \textbf{87.75}	&  87.62	& 87.36           \\
          MP &86.49 	& 87.24	&\textbf{87.62}	& 85.08        &86.99 	& \textbf{87.62}	&  86.84	& 87.14\\
        MR &\textbf{87.29} 	& \textbf{87.29}	&86.85	& 86.73              &87.25 	& 86.85	&\textbf{87.66}	& 86.37 \\
         Macro F1  &86.87 	& 87.06	&\textbf{87.22}	& 85.79             &87.11 & 87.22	&\textbf{87.23}	& 86.73                \\
        \bottomrule
	\end{tabular}}
	\end{threeparttable}
	 \end{center}
\end{table}

\subsection{Evaluation on the diagnosis of COVID-19}

 We then evaluate our method by diagnosing the new coronavirus disease 2019 (COVID-19) based on chest X-ray images.

\textbf{Dataset}.  The  used dataset is collected from Kaggle, consisting of the   COVID-19 Radiography Database  and the dataset of RSNA Pneumonia Detection Challenge.
Based on the collected dataset\footnote{The up-to-date dataset is available: https://github.com/Vanint/COVID-DA.}, we randomly choose part of normal cases and all typical pneumonia cases to make up the source domain, and use the rest of normal cases and all COVID-19 cases as the target domain~\cite{zhang2020covid}.
The statistics of two domains are summarized in Table~\ref{tab:data_covid}.
Since this dataset is clean, we construct its corrupted counterpart. Following~\cite{ref_mentornet,ref_tcl}, we change the label of each image in the source domain uniformly to another class with probability $\rho_{\text{noise}}\small{=}0.1$.

\begin{table}[h]
  \caption{ {Statistics of the dataset, where pneumonia serves as the source domain and COVID-19 serves as the target domain.}}
  \vspace{-0.1in}
  \centering
  \begin{threeparttable}
  \renewcommand\arraystretch{0.7}
  \renewcommand{\tabcolsep}{2.5pt}
  \resizebox{0.46\textwidth}{!}{

    \begin{tabular}[width=\textwidth]{cccccccccc}
    \toprule
    \multirow{2}{*}{Set} &\multirow{2}{*}{Domain}&
    \multicolumn{3}{c}{Categories}&
    \multirow{2}{*}{\#Total}\cr
    \cmidrule(lr){3-5}
    & &  \#Normal & \#Pneumonia & \#COVID-19& & \cr
    \midrule
    \multirow{2}{*}{Training} & Pneumonia  & 5,613 & 2,298 & 0 & 7,911\cr
     & COVID-19 & 2,541 & 0 & 176 & 2,717\cr
    \midrule
    Test& COVID-19 & 885 & 0 & 43 & 928\cr
    \bottomrule
    \end{tabular}
  }
    \end{threeparttable}
    \label{tab:data_covid}
\end{table}

\textbf{Results}.
As shown in Table~\ref{performance_covid}, CoUDA outperforms all baselines significantly. This demonstrates that CoUDA can be applied to X-ray images in general and to COVID-19 X-ray based diagnosis. Since the outbreak of COVID-19 has already infected millions of people but extensive annotations of COVID-19 are inaccessible now, such a result verifies the  importance of CoUDA for developing deep learning based diagnosis methods for COVID-19. Moreover, we hope our preliminary results on the limited
amount of COVID-19 data can inspire more research on UDA for COVID-19 in the future.

 \begin{table}[h]
	\caption{ {Comparisons on COVID-19 diagnosis based on chest X-ray iamges, where noise rate is set to 0.1.}}
	\label{performance_covid}
 \begin{center}
 \begin{threeparttable}
 	\begin{tabular}{lcccc}\toprule
        Methods
        & Acc (\%) & MP & MR  & Macro F1  \cr
        \midrule
         MobileNet~[63]  &58.50	 	& 47.77	& 31.31	& 37.07      \\
         MentorNet~[50]  &78.35	 	& 48.44	& 41.53	& 44.25       \\
         Co-teaching~[21]  & 91.27 &63.43	& 76.62	& 67.29            \\	
         NAL~[30]&  94.29 	& 69.80	& 77.09	& 72.76               \\
         \midrule
         DDC~[18] & 78.95 & 49.03	& 44.80 		& 45.26           \\
         DANN~[17] & 87.18 	& 51.58	& 55.55 	& 51.09               \\
         MCD~[24] & 96.77 	& 84.31	& 75.07	& 78.79                   \\
         CLAN~[20] &95.58	& 75.04	& 71.13	& 72.90                  \\
         TCL~[54] & 97.20	& 86.87	& 78.62	& 82.16                  \\
          \midrule
          CoUDA  & \textbf{98.06} 	& \textbf{94.33}	& \textbf{82.39} 	& \textbf{87.33}          \\
        \bottomrule

	\end{tabular}
	
	 \end{threeparttable}
	 \end{center}
	 \label{tab:covid19_experi}
\end{table}

\textbf{Evaluation on diverse noise rates}.
 {The above experiment verifies CoUDA for diagnosing COVID-19 under  noise rate $0.1$. This subsection further evaluates CoUDA under the noise rates of 0.2 and 0.4. Table~\ref{tab:further_noisy_rates} shows that CoUDA  consistently outperforms two state-of-the-art baselines (\ie CLAN and TCL) under different noise rates. This further verifies the superiority of CoUDA when dealing with label noise.}

\begin{table}[h]
	\caption{ {Comparisons on the  COVID-19 dataset under different noise rates.}}
	\label{performance}
 \begin{center}
 \begin{threeparttable}
 	\begin{tabular}{lcccc}\toprule
        \multirow{2}{*}{Methods}&\multicolumn{4}{c}{noise rate = 0.2}\cr\cmidrule{2-5}
        & Acc (\%) & MP & MR  & Macro F1  \cr
        \midrule
         CLAN~[20] &88.15	& 59.27	& 72.77	& 61.95                  \\
         TCL~[54] & 96.34	& 83.00	& 68.21	& 73.29                  \\
          CoUDA   & \textbf{97.31} 	& \textbf{86.00}	& \textbf{81.99} 	& \textbf{83.86}          \\
          \midrule
        \multirow{2}{*}{Methods}&\multicolumn{4}{c}{noise rate = 0.4}\cr\cmidrule{2-5}
        & Acc (\%) & MP & MR  & Macro F1  \cr
        \midrule
         CLAN~[20] &85.67	& 51.64 & 53.77 & 51.47                  \\
         TCL~[54] & 89.01	& 58.00	& 65.47	& 59.53                 \\
          CoUDA & \textbf{95.69} 	& \textbf{76.08}	& \textbf{67.87} 	& \textbf{71.10}          \\
        \bottomrule

	\end{tabular}
	 \end{threeparttable}
	 \end{center}
	 \label{tab:further_noisy_rates}
\end{table}

   \begin{table*}[ht]
   \vspace{0.1in}
	 \caption{Comparisons on general images  on the Office-31 dataset under the fixed noise rate 0.1, where ``avg." denotes the average result of six transfer tasks.} \vspace{0.1in}
	\label{performance_office}
     \begin{center}
     \begin{threeparttable}
       \renewcommand\arraystretch{1}
        \renewcommand{\tabcolsep}{2.6pt}
    \resizebox{1\textwidth}{!}{
     	\begin{tabular}{lccccccc|ccccccc}
     	\toprule
             \multirow{2}{*}{Methods}&\multicolumn{7}{c|}{Acc (\%)}&\multicolumn{7}{c}{Macro F1-measure}\cr\cmidrule{2-8}\cmidrule{9-15}
            & A$\rightarrow$ W & D$\rightarrow$ W & W$\rightarrow$D & A$\rightarrow$ D    &D$\rightarrow$ A  & W$\rightarrow$ A & avg. & A$\rightarrow$ W & D$\rightarrow$ W & W$\rightarrow$D & A$\rightarrow$ D    &D$\rightarrow$ A  & W$\rightarrow$ A & avg. \cr
            \midrule
             MobileNet~\cite{ref_Mobilenet}    &39.85 	 	& 77.51 	& 62.95	& 40.80          &30.63	 	& 33.91	& 47.61 &37.33	 	& 75.88	& 62.74	& 32.83         &28.39	 	& 32.11& 45.38\\
             MentorNet~\cite{ref_mentornet}  &34.98	 	& 69.60	& 68.40	& 44.40        &38.07	 	& 38.25 & 48.95 &35.62	 	& 64.95 	& 63.47	& 42.04      & 36.00 	& 36.00 & 46.34\\
             Co-teaching~\cite{ref_coteaching}   & 42.96 &70.38	& 78.89 & 47.01            &40.24	 	&40.87 & 53.39 & 43.59 &69.26	& 69.56	& 41.18            &38.02	 	& 39.06	& 50.11\\	
             NAL~\cite{ref_noiselayer}&45.34 	& 74.22	& 72.51	&50.60          &29.27	 	& 37.05	 &51.50 &45.05 	& 72.24	& 65.81	& 45.14              &28.09	 	&33.56	 & 48.32\\
             \midrule
             DDC~\cite{ref_DDC} &52.20 	& 80.71	& 84.40	&50.80          &38.12	 	& 36.75	 &57.16 & 45.80 & 78.70	& 79.84 		& 44.88	            &35.83	 	&36.01  & 53.51\\
             DANN~\cite{ref_DANN} & 59.78 	& 84.64	& 86.85 	& 50.80              &43.38	 	& 42.92 & 61.39 & 44.90 	& 83.31	& 78.88 	& 50.06              &40.52	 	& 42.11 & 56.63\\
             MCD~\cite{ref_mcd} &43.88 	& 85.52	& 88.50	& 56.97	 	& 42.65 	 	& 45.56	& 60.51 &40.32 	& 84.37	& 80.13	& 47.17        &41.09	 	& 44.81	  &   56.32\\
             CLAN~\cite{luo2019taking} &42.96	& 85.29 &	88.05	& 60.96& 	39.78& 	44.83& 	60.31 & 41.24 &	84.63 &	79.58 &	52.95 &	37.53 &	42.62 &	56.43\\
             TCL~\cite{ref_tcl}  &46.62 	& 85.56	& 85.90	& 56.18                  &41.69	 	& 44.20	& 60.02 &46.04 	& 83.00	& 80.90	& 47.81                  &41.72	 	& 40.29	& 56.68\\
             \midrule
             CoUDA  & \textbf{63.07} 	& \textbf{85.92}	& \textbf{88.84} 	& \textbf{62.15}     &    \textbf{45.74} 	& \textbf{46.24}  & \textbf{65.33}	 & \textbf{61.77} 	& \textbf{85.79}	& \textbf{83.62} 	& \textbf{57.38}     &    \textbf{43.66} 	& \textbf{46.02}  & \textbf{63.04}   \\
            \bottomrule
    	\end{tabular} }
    	\end{threeparttable}
    	\end{center}
    	\vspace{0.1in}
    	 \end{table*}

 {\subsection{Application to General Images}}

\textbf{Dataset}:  Office-31~\cite{ref_office} is a standard dataset for domain adaptation. It consists of 4652 images with 31 classes in 3 distinct domains: amazon (A), webcam (W) and dslr (D), which are collected from amazon.com, web cameras, and digital SLR cameras, respectively. By permuting the 3 domains, we obtain 6 unsupervised transfer tasks. According to the standard setting of Office-31~\cite{ref_office}, the source domain has 8 labeled data per class for webcam/dslr and 20 for amazon. Since this dataset is almost clean and class balanced, we create its corrupted counterpart. Following~\cite{ref_mentornet,ref_tcl}, we change the label of each image uniformly to a random class with probability $p_{\text{noise}}=0.1$. For the class imbalance, we randomly select each class with probability $p_{\text{class}}=0.5$ to reduce half data.  Like~\cite{ref_tcl}, we use the noisy domain as the source domain and the clean domain as the target domain. Note that this task is very challenging, since labeled source data is very limited and also suffers class corruption.

\textbf{Results}:
 {We report the result on Office-31 in Table~\ref{performance_office}.
Promising result verifies the effectiveness of  the proposed method in handling general image tasks. In addition, the result also reveals several observations.
(1) Domain discrepancy and label noise limits the performance of standard deep neural networks on the target domain (MobileNet).
(2) Ignoring the class imbalance issue affects the effectiveness of handling label noise (MentorNet, Co-teaching and TCL).
(3) Domain adaptation enhances the model performance on the target domain, but most  existing methods cannot handle the issue of label noise in medical image diagnosis (DDC, DANN, MCD, CLAN).
(4) Cooperation is helpful to handle the domain discrepancy and label noise simultaneously  (CoUDA).
}

\textbf{Estimation of Noise Transition Matrix}:
We next evaluate the estimated  noise transition matrix, learned by the noise co-adaptation layer.
As mentioned above,
we construct the corrupted Office-31 by changing the label of each image uniformly to a random class with probability $0.1$. Hence, the  true noise transition matrix is:
 $\Big{[}0.9*\mathbb{I}_{(i=j)}+\frac{0.1}{31}*\mathbb{I}_{(i\neq j)}\Big{]}$,
where $i\small{\in}\{0,..,30\}$ is the true label and $j\small{\in}\{0,..,30\}$ is the changed noisy label. Parameter initialization of the noise co-adaptation layer is based on the method in Section V.B.1 with $\epsilon\small{=}0.8$. Hence, there is a gap between the true transition matrix and the initial estimated one.

\begin{figure}[t]
    \vspace{0.1in}
    \centerline{\includegraphics[width=7.3cm]{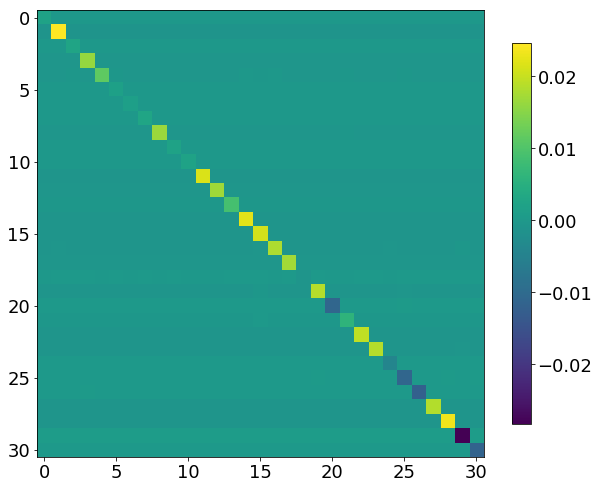}}
    \vspace{0.1in}
    \caption{An empirical example of the difference between the true noise transition matrix and the estimated one by the noise co-adaptation layer on $W\rightarrow D$ of Office-31.}
    \label{difference}
\end{figure}

Fig.~\ref{difference} presents the difference between the true noise transition matrix and the estimated one by the noise co-adaptation layer on the setting of $W\rightarrow D$. As is shown in Fig.~\ref{difference}, the noise co-adaptation layer   approximates the true noise transition matrix
within an acceptable error level. This result not only verifies the effectiveness of the proposed noise co-adaptation layer, but also explains the state-of-the-art performance of CoUDA in previous experiments.

\textbf{Evaluation on diverse noise rates}.
In the previous experiments, we have demonstrated the proposed CoUDA on Office-31 with the fixed label noise rate $0.1$. Here, we further evaluate our method on Office-31 corrupted by the noise rates of 0.2 and 0.4. As shown in Table~\ref{noise_diff},  our proposed method deals with label noise consistently better than other state-of-the-art unsupervised domain adaptation methods (\ie CLAN and TCL).\\

\begin{table}[t]
	\caption{Comparisons on Office-31 under different noise rates in terms of Macro F1 measure. ``avg." denotes the average result of six transfer tasks.}
	\label{noise_diff}
	\vspace{0.1in}
 \begin{center}
      \begin{threeparttable}
      \renewcommand{\tabcolsep}{2.6pt}
    \resizebox{0.49\textwidth}{!}{
 	\begin{tabular}{lccccccc}\toprule
        \multirow{2}{*}{Methods}&\multicolumn{7}{c}{noise rate = 0.2}  \cr\cmidrule{2-8}
        & A$>$W & D$>$W & W$>$D & A$>$D    &D$>$A  & W$>$A &avg. \cr
        \midrule
          {CLAN~\cite{luo2019taking}}  & {43.30} 	&  {69.52}	&  {70.68}	&  {44.95}   & {29.13}	 	&  {31.58}	&  {48.19}  \\
         TCL~\cite{ref_tcl}  &38.94 	& 68.56	& 64.54	& 30.68   &23.26	 	& 23.21	& 41.53   \\
         CoUDA  & \textbf{53.75} 	& \textbf{75.86}	& \textbf{72.77} 	& \textbf{48.28}     &    \textbf{36.71} 	& \textbf{33.81}  & \textbf{53.49} 		   \\
        \midrule
        \multirow{2}{*}{Methods}&\multicolumn{7}{c}{noise rate = 0.4}\cr\cmidrule{2-8}
        & A$>$W & D$>$W & W$>$D & A$>$D    &D$>$A  & W$>$A & avg.\cr
        \midrule
          {CLAN~\cite{luo2019taking}}  & {31.46} 	&  {41.37}	&  {59.34}	&  {37.69}   & {22.02}	 	&  {28.10}	&  {36.66}  \\
         TCL~\cite{ref_tcl}    &30.04 	&51.01	& 50.12	& 16.34     &  15.07	 	& 12.24	& 29.14\\
         CoUDA  & \textbf{45.03} 	& \textbf{65.76}	& \textbf{63.52} 	& \textbf{38.15}     &    \textbf{32.66} 	& \textbf{30.93}  & \textbf{46.01}		   \\
        \bottomrule
	\end{tabular} }
	\end{threeparttable}
	 \end{center} \vspace{0.1in}
\end{table}

\section{Conclusion}
In this paper, we have proposed a general  Collaborative Unsupervised Domain Adaptation method. Unlike previous UDA methods that treat all data equally or assume data with clean labels, our method collaboratively eliminates the domain discrepancy with more focuses on hard-to-transfer samples, and overcome label noise simultaneously.
Also, we theoretically analyze the generalization error of the proposed method.
Extensive experiments on real-world medical and general images demonstrate the effectiveness and generalization of the proposed method.
Since there is an urgent need for robust deep models for medical image diagnosis, our method is of great clinical significance in practice.
Moreover, considering that UDA with noise labels is not specific to the medical domain, we expect that our method will be applied to more general image classification tasks, such as in crowdsourcing and financial domains.

\bibliographystyle{IEEEtran}
\bibliography{CoUDA}

\begin{thebibliography}{10}
\providecommand{\url}[1]{#1}
\csname url@samestyle\endcsname
\providecommand{\newblock}{\relax}
\providecommand{\bibinfo}[2]{#2}
\providecommand{\BIBentrySTDinterwordspacing}{\spaceskip=0pt\relax}
\providecommand{\BIBentryALTinterwordstretchfactor}{4}
\providecommand{\BIBentryALTinterwordspacing}{\spaceskip=\fontdimen2\font plus
\BIBentryALTinterwordstretchfactor\fontdimen3\font minus
  \fontdimen4\font\relax}
\providecommand{\BIBforeignlanguage}[2]{{%
\expandafter\ifx\csname l@#1\endcsname\relax
\typeout{** WARNING: IEEEtran.bst: No hyphenation pattern has been}%
\typeout{** loaded for the language `#1'. Using the pattern for}%
\typeout{** the default language instead.}%
\else
\language=\csname l@#1\endcsname
\fi
#2}}
\providecommand{\BIBdecl}{\relax}
\BIBdecl

\bibitem{jia2019deep}
K.~Jia, J.~Lin, M.~Tan, and D.~Tao, ``Deep multi-view learning using
  neuron-wise correlation-maximizing regularizers,'' \emph{IEEE TIP}, 2019.

\bibitem{chen2020scripted}
Q.~Chen, Q.~Wu, J.~Chen, Q.~Wu, A.~Van Den~Hengel, and M.~Tan, ``Scripted video
  generation with a bottom-up generative adversarial network,'' \emph{IEEE
  TIP}, 2020.

\bibitem{zhang2019whole}
Y.~Zhang, H.~Chen, Y.~Wei, P.~Zhao \emph{et~al.}, ``From whole slide imaging to
  microscopy: Deep microscopy adaptation network for histopathology cancer
  image classification,'' in \emph{MICCAI}, 2019.

\bibitem{cao2019multi}
J.~Cao, L.~Mo, Y.~Zhang, K.~Jia, C.~Shen, and M.~Tan, ``Multi-marginal
  wasserstein gan,'' in \emph{NeurIPS}, 2019.

\bibitem{yao2018deep}
J.~Yao, J.~Wang, I.~W. Tsang \emph{et~al.}, ``Deep learning from noisy image
  labels with quality embedding,'' \emph{IEEE TIP}, 2018.

\bibitem{chan2015pcanet}
T.-H. Chan, K.~Jia, S.~Gao, J.~Lu, Z.~Zeng, and Y.~Ma, ``Pcanet: A simple deep
  learning baseline for image classification?'' \emph{IEEE TIP}, 2015.

\bibitem{lim2017enhanced}
B.~Lim, S.~Son, H.~Kim, S.~Nah, and K.~Mu~Lee, ``Enhanced deep residual
  networks for single image super-resolution,'' in \emph{CVPR Workshop}, 2017.

\bibitem{ref_partial1}
Z.~Cao, K.~You, M.~Long, J.~Wang, and Q.~Yang, ``Learning to transfer examples
  for partial domain adaptation,'' in \emph{CVPR}, 2019.

\bibitem{dubey2015local}
S.~R. Dubey, S.~K. Singh, and R.~K. Singh, ``Local wavelet pattern: a new
  feature descriptor for image retrieval in medical ct databases,'' \emph{IEEE
  TIP}, 2015.

\bibitem{zhou2019high}
S.~Zhou, D.~Nie, E.~Adeli, J.~Yin, J.~Lian, and D.~Shen, ``High-resolution
  encoder--decoder networks for low-contrast medical image segmentation,''
  \emph{IEEE TIP}, 2019.

\bibitem{khadidos2017weighted}
A.~Khadidos, V.~Sanchez, and C.-T. Li, ``Weighted level set evolution based on
  local edge features for medical image segmentation,'' \emph{IEEE TIP}, 2017.

\bibitem{zhang2020covid}
Y.~Zhang, S.~Niu, Z.~Qiu, Y.~Wei, P.~Zhao, J.~Yao, J.~Huang, Q.~Wu, and M.~Tan,
  ``Covid-da: Deep domain adaptation from typical pneumonia to covid-19,''
  \emph{arXiv}, 2020.

\bibitem{ref_survey2}
G.~Litjens, T.~Kooi, B.~E. Bejnordi \emph{et~al.}, ``A survey on deep learning
  in medical image analysis,'' \emph{Medical image analysis}, 2017.

\bibitem{ref_survey}
F.~Xing, Y.~Xie, H.~Su, F.~Liu, and L.~Yang, ``Deep learning in microscopy
  image analysis: A survey,'' \emph{TNNLS}, 2017.

\bibitem{ref_domain_miccai}
T.~Heimann, P.~Mountney, M.~John, and R.~Ionasec, ``Learning without labeling:
  Domain adaptation for ultrasound transducer localization,'' in \emph{MICCAI},
  2013.

\bibitem{ref_transfer}
S.~J. Pan and Q.~Yang, ``A survey on transfer learning,'' \emph{TKDE}, 2009.

\bibitem{zhang2019collaborative}
Y.~Zhang, Y.~Wei, P.~Zhao, S.~Niu, Q.~Wu, M.~Tan, and J.~Huang, ``Collaborative
  unsupervised domain adaptation for medical image diagnosis,'' \emph{Medical
  Imaging meets NeurIPS}, 2019.

\bibitem{ref_DANN}
Y.~Ganin and V.~Lempitsky, ``Unsupervised domain adaptation by
  backpropagation,'' in \emph{ICML}, 2015.

\bibitem{ref_DDC}
E.~Tzeng, J.~Hoffman, N.~Zhang, K.~Saenko, and T.~Darrell, ``Deep domain
  confusion: Maximizing for domain invariance,'' \emph{arXiv}, 2014.

\bibitem{ref_ADDA}
E.~Tzeng, J.~Hoffman, K.~Saenko, and T.~Darrell, ``Adversarial discriminative
  domain adaptation,'' in \emph{CVPR}, 2017.

\bibitem{luo2019taking}
Y.~Luo, L.~Zheng, T.~Guan, J.~Yu, and Y.~Yang, ``Taking a closer look at domain
  shift: Category-level adversaries for semantics consistent domain
  adaptation,'' in \emph{CVPR}, 2019, pp. 2507--2516.

\bibitem{ref_coteaching}
B.~Han, Q.~Yao, X.~Yu, G.~Niu, M.~Xu, W.~Hu \emph{et~al.}, ``Co-teaching:
  Robust training of deep neural networks with extremely noisy labels,'' in
  \emph{NeurIPS}, 2018.

\bibitem{ref_coteaching1}
X.~Yu, B.~Han, J.~Yao \emph{et~al.}, ``How does disagreement help
  generalization against label corruption?'' in \emph{ICML}, 2019.

\bibitem{ref_Rademacher_Bartlett}
P.~L. Bartlett and S.~Mendelson, ``Rademacher and gaussian complexities: Risk
  bounds and structural results,'' \emph{JMLR}, 2002.

\bibitem{Hoi_2006}
S.~C. Hoi, R.~Jin, J.~Zhu \emph{et~al.}, ``Batch mode active learning and its
  application to medical image classification,'' in \emph{ICML}, 2006.

\bibitem{ref_adaptation1}
C.~Becker, C.~M. Christoudias \emph{et~al.}, ``Domain adaptation for microscopy
  imaging,'' \emph{IEEE TMI}, 2014.

\bibitem{ref_adaptation2}
R.~Berm{\'u}dez-Chac{\'o}n, C.~Becker \emph{et~al.}, ``Scalable unsupervised
  domain adaptation for electron microscopy,'' in \emph{MICCAI}, 2016.

\bibitem{ref_res}
M.~Long, H.~Zhu, J.~Wang, and M.~I. Jordan, ``Unsupervised domain adaptation
  with residual transfer networks,'' in \emph{NeurIPS}, 2016.

\bibitem{ref_cada}
M.~Long, Z.~Cao, J.~Wang, and M.~I. Jordan, ``Conditional adversarial domain
  adaptation,'' in \emph{NeurIPS}, 2018.

\bibitem{transferable_attention}
X.~Wang, L.~Li, W.~Ye, M.~Long, and J.~Wang, ``Transferable attention for
  domain adaptation,'' in \emph{AAAI}, 2019.

\bibitem{ref_mcd}
K.~Saito, K.~Watanabe, Y.~Ushiku \emph{et~al.}, ``Maximum classifier
  discrepancy for unsupervised domain adaptation,'' in \emph{CVPR}, 2018.

\bibitem{ref_SPL}
M.~P. Kumar, B.~Packer, and D.~Koller, ``Self-paced learning for latent
  variable models,'' in \emph{NeurIPS}, 2010.

\bibitem{ref_mentornet}
L.~Jiang, Z.~Zhou, T.~Leung, L.-J. Li, and L.~Fei-Fei, ``Mentornet: Learning
  data-driven curriculum for very deep neural networks on corrupted labels,''
  in \emph{ICML}, 2018.

\bibitem{Natarajan2013learning}
N.~Natarajan, I.~S. Dhillon, P.~K. Ravikumar, and A.~Tewari, ``Learning with
  noisy labels,'' in \emph{NeurIPS}, 2013.

\bibitem{ref_Patrini}
G.~Patrini, A.~Rozza, A.~Krishna~Menon, R.~Nock, and L.~Qu, ``Making deep
  neural networks robust to label noise: A loss correction approach,'' in
  \emph{CVPR}, 2017.

\bibitem{ref_Sukhbaatar2014}
S.~Sukhbaatar and R.~Fergus, ``Learning from noisy labels with deep neural
  networks,'' \emph{arXiv}, 2014.

\bibitem{ref_noiselayer}
J.~Goldberger and E.~Ben-Reuven, ``Training deep neural-networks using a noise
  adaptation layer,'' in \emph{ICLR}, 2017.

\bibitem{dimensionality}
X.~Ma, Y.~Wang, M.~E. Houle, S.~Zhou, S.~M. Erfani \emph{et~al.},
  ``Dimensionality-driven learning with noisy labels,'' in \emph{ICML}, 2018.

\bibitem{bootstrap1}
E.~Arazo, D.~Ortego, P.~Albert, N.~E. O'Connor \emph{et~al.}, ``Unsupervised
  label noise modeling and loss correction,'' in \emph{ICML}, 2019.

\bibitem{ref_bootstrap}
S.~Reed, H.~Lee, D.~Anguelov \emph{et~al.}, ``Training deep neural networks on
  noisy labels with bootstrapping,'' \emph{arXiv}, 2014.

\bibitem{ref_tcl}
Y.~Shu, Z.~Cao, M.~Long, and J.~Wang, ``Transferable curriculum for
  weakly-supervised domain adaptation,'' in \emph{AAAI}, 2019.

\bibitem{Blum1998}
A.~Blum and T.~Mitchell, ``Combining labeled and unlabeled data with
  co-training,'' in \emph{COLT}, 1998.

\bibitem{Kumar2011}
A.~Kumar and H.~Daum{\'e}, ``A co-training approach for multi-view spectral
  clustering,'' in \emph{ICML}, 2011.

\bibitem{Chen2011train}
M.~Chen, K.~Weinberger, and J.~Blitzer, ``Co-training for domain adaptation,''
  in \emph{NeurIPS}, 2011.

\bibitem{Zhou2005}
Z.-H. Zhou and M.~Li, ``Semi-supervised regression with co-training.'' in
  \emph{IJCAI}, 2005.

\bibitem{multiagent}
L.~Panait and S.~Luke, ``Cooperative multi-agent learning: The state of the
  art,'' \emph{Autonomous Agents and Multi-agent Systems}, 2005.

\bibitem{ref_coregularized}
A.~Kumar, P.~Sattigeri, K.~Wadhawan, L.~Karlinsky, R.~Feris, B.~Freeman, and
  G.~Wornell, ``Co-regularized alignment for unsupervised domain adaptation,''
  in \emph{NeurIPS}, 2018.

\bibitem{tri_training}
Z.-H. Zhou and M.~Li, ``Tri-training: Exploiting unlabeled data using three
  classifiers,'' \emph{TKDE}, 2005.

\bibitem{tri_training1}
K.~Saito, Y.~Ushiku, and T.~Harada, ``Asymmetric tri-training for unsupervised
  domain adaptation,'' in \emph{ICML}, 2017.

\bibitem{ref_GAN}
I.~Goodfellow, J.~Pouget-Abadie, M.~Mirza, B.~Xu, D.~Warde-Farley, S.~Ozair,
  A.~Courville, and Y.~Bengio, ``Generative adversarial nets,'' in
  \emph{NeurIPS}, 2014.

\bibitem{ref_LSGAN}
X.~Mao, Q.~Li, H.~Xie, R.~Y. Lau, Z.~Wang, and S.~Paul~Smolley, ``Least squares
  generative adversarial networks,'' in \emph{ICCV}, 2017.

\bibitem{rethinking}
C.~Zhang, S.~Bengio, M.~Hardt \emph{et~al.}, ``Understanding deep learning
  requires rethinking generalization,'' in \emph{ICLR}, 2017.

\bibitem{ref_Focal}
T.-Y. Lin, P.~Goyal, R.~Girshick, K.~He, and P.~Doll{\'a}r, ``Focal loss for
  dense object detection,'' in \emph{ICCV}, 2017.

\bibitem{ref_ensemble}
Z.-H. Zhou, \emph{Ensemble methods: foundations and algorithms}.\hskip 1em plus
  0.5em minus 0.4em\relax Chapman and Hall/CRC, 2012.

\bibitem{ref_partial}
Z.~Cao, M.~Long, J.~Wang, and M.~I. Jordan, ``Partial transfer learning with
  selective adversarial networks,'' in \emph{CVPR}, 2018.

\bibitem{Yu2018learning}
X.~Yu, T.~Liu, M.~Gong, and D.~Tao, ``Learning with biased complementary
  labels,'' in \emph{ECCV}, 2018.

\bibitem{Koltchinskii2002empirical}
V.~Koltchinskii, D.~Panchenko \emph{et~al.}, ``Empirical margin distributions
  and bounding the generalization error of combined classifiers,'' \emph{The
  Annals of Statistics}, 2002.

\bibitem{Chazelle2001the}
B.~Chazelle, \emph{The discrepancy method: randomness and complexity}.\hskip
  1em plus 0.5em minus 0.4em\relax Cambridge University Press, 2001.

\bibitem{Mansour2009domain}
Y.~Mansour, M.~Mohri, and A.~Rostamizadeh, ``Domain adaptation: Learning bounds
  and algorithms,'' in \emph{COLT}, 2009.

\bibitem{ref_Mobilenet}
M.~Sandler, A.~Howard, M.~Zhu, A.~Zhmoginov, and L.-C. Chen, ``Mobilenetv2:
  Inverted residuals and others,'' in \emph{CVPR}, 2018.

\bibitem{ref_imagenet}
O.~Russakovsky, J.~Deng \emph{et~al.}, ``Imagenet large scale visual
  recognition challenge,'' \emph{IJCV}, 2015.

\bibitem{metric}
H.~Koppula \emph{et~al.}, ``Learning spatio-temporal structure from rgb-d
  videos for human activity detection and anticipation,'' in \emph{ICML}, 2013.

\bibitem{zhang2018online}
Y.~Zhang, P.~Zhao, J.~Cao, W.~Ma, J.~Huang, Q.~Wu, and M.~Tan, ``Online
  adaptive asymmetric active learning for budgeted imbalanced data,'' in
  \emph{SIGKDD}, 2018, pp. 2768--2777.

\bibitem{ref_tsne}
L.~v.~d. Maaten \emph{et~al.}, ``Visualizing data using t-sne,'' \emph{JLMR},
  2008.

\bibitem{ref_office}
K.~Saenko, B.~Kulis, M.~Fritz, and T.~Darrell, ``Adapting visual category
  models to new domains,'' in \emph{ECCV}, 2010.

\end{thebibliography}


\begin{thebibliography}{10}
\providecommand{\url}[1]{#1}
\csname url@samestyle\endcsname
\providecommand{\newblock}{\relax}
\providecommand{\bibinfo}[2]{#2}
\providecommand{\BIBentrySTDinterwordspacing}{\spaceskip=0pt\relax}
\providecommand{\BIBentryALTinterwordstretchfactor}{4}
\providecommand{\BIBentryALTinterwordspacing}{\spaceskip=\fontdimen2\font plus
\BIBentryALTinterwordstretchfactor\fontdimen3\font minus
  \fontdimen4\font\relax}
\providecommand{\BIBforeignlanguage}[2]{{%
\expandafter\ifx\csname l@#1\endcsname\relax
\typeout{** WARNING: IEEEtran.bst: No hyphenation pattern has been}%
\typeout{** loaded for the language `#1'. Using the pattern for}%
\typeout{** the default language instead.}%
\else
\language=\csname l@#1\endcsname
\fi
#2}}
\providecommand{\BIBdecl}{\relax}
\BIBdecl

\bibitem{Chazelle2001the}
B.~Chazelle, \emph{The discrepancy method: randomness and complexity}.\hskip
  1em plus 0.5em minus 0.4em\relax Cambridge University Press, 2001.

\bibitem{Mansour2009domain}
Y.~Mansour, M.~Mohri, and A.~Rostamizadeh, ``Domain adaptation: Learning bounds
  and algorithms,'' in \emph{COLT}, 2009.

\bibitem{ref_Rademacher_Bartlett}
P.~L. Bartlett and S.~Mendelson, ``Rademacher and gaussian complexities: Risk
  bounds and structural results,'' \emph{JMLR}, 2002.

\bibitem{Koltchinskii2002empirical}
V.~Koltchinskii, D.~Panchenko \emph{et~al.}, ``Empirical margin distributions
  and bounding the generalization error of combined classifiers,'' \emph{The
  Annals of Statistics}, 2002.

\bibitem{Ledoux2013}
M.~Ledoux and M.~Talagrand, \emph{Probability in Banach Spaces: isoperimetry
  and processes}.\hskip 1em plus 0.5em minus 0.4em\relax Springer Science \&
  Business Media, 2013.

\bibitem{Yu2018learning}
X.~Yu, T.~Liu, M.~Gong, and D.~Tao, ``Learning with biased complementary
  labels,'' in \emph{ECCV}, 2018.

\bibitem{Natarajan2013learning}
N.~Natarajan, I.~S. Dhillon, P.~K. Ravikumar, and A.~Tewari, ``Learning with
  noisy labels,'' in \emph{NeurIPS}, 2013.

\bibitem{Wan2013Regularization}
L.~Wan, M.~Zeiler, S.~Zhang, Y.~Le~Cun, and R.~Fergus, ``Regularization of
  neural networks using dropconnect,'' in \emph{ICML}, 2013.

\bibitem{Zhai2018}
K.~Zhai and H.~Wang, ``Adaptive dropout with rademacher complexity
  regularization,'' in \emph{ICLR}, 2018.

\bibitem{ref_Mobilenet}
M.~Sandler, A.~Howard, M.~Zhu, A.~Zhmoginov, and L.-C. Chen, ``Mobilenetv2:
  Inverted residuals and others,'' in \emph{CVPR}, 2018.

\bibitem{ref_noiselayer}
J.~Goldberger and E.~Ben-Reuven, ``Training deep neural-networks using a noise
  adaptation layer,'' in \emph{ICLR}, 2017.

\bibitem{ref_Sukhbaatar2014}
S.~Sukhbaatar and R.~Fergus, ``Learning from noisy labels with deep neural
  networks,'' \emph{arXiv}, 2014.

\bibitem{ref_ADDA}
E.~Tzeng, J.~Hoffman, K.~Saenko, and T.~Darrell, ``Adversarial discriminative
  domain adaptation,'' in \emph{CVPR}, 2017.

\bibitem{ref_LSGAN}
X.~Mao, Q.~Li, H.~Xie, R.~Y. Lau, Z.~Wang, and S.~Paul~Smolley, ``Least squares
  generative adversarial networks,'' in \emph{ICCV}, 2017.

\bibitem{ref_GAN}
I.~Goodfellow, J.~Pouget-Abadie, M.~Mirza, B.~Xu, D.~Warde-Farley, S.~Ozair,
  A.~Courville, and Y.~Bengio, ``Generative adversarial nets,'' in
  \emph{NeurIPS}, 2014.

\bibitem{ref_entropymin}
J.-F. Mangin, ``Entropy minimization for automatic correction of intensity
  nonuniformity,'' in \emph{Workshop on MMBIA}, 2000.

\bibitem{ref_tcl}
Y.~Shu, Z.~Cao, M.~Long, and J.~Wang, ``Transferable curriculum for
  weakly-supervised domain adaptation,'' in \emph{AAAI}, 2019.

\bibitem{ref_ensemble}
Z.-H. Zhou, \emph{Ensemble methods: foundations and algorithms}.\hskip 1em plus
  0.5em minus 0.4em\relax Chapman and Hall/CRC, 2012.

\bibitem{ref_DANN}
Y.~Ganin and V.~Lempitsky, ``Unsupervised domain adaptation by
  backpropagation,'' in \emph{ICML}, 2015.

\bibitem{zhang2019whole}
Y.~Zhang, H.~Chen, Y.~Wei, P.~Zhao \emph{et~al.}, ``From whole slide imaging to
  microscopy: Deep microscopy adaptation network for histopathology cancer
  image classification,'' in \emph{MICCAI}, 2019.

\bibitem{zhang2019collaborative}
Y.~Zhang, Y.~Wei, P.~Zhao, S.~Niu, Q.~Wu, M.~Tan, and J.~Huang, ``Collaborative
  unsupervised domain adaptation for medical image diagnosis,'' \emph{Medical
  Imaging meets NeurIPS}, 2019.

\bibitem{ref_partial}
Z.~Cao, M.~Long, J.~Wang, and M.~I. Jordan, ``Partial transfer learning with
  selective adversarial networks,'' in \emph{CVPR}, 2018.

\end{thebibliography}

\clearpage
\begin{IEEEbiography}[{\includegraphics[width=1.2in,height=1.25in,clip,keepaspectratio]{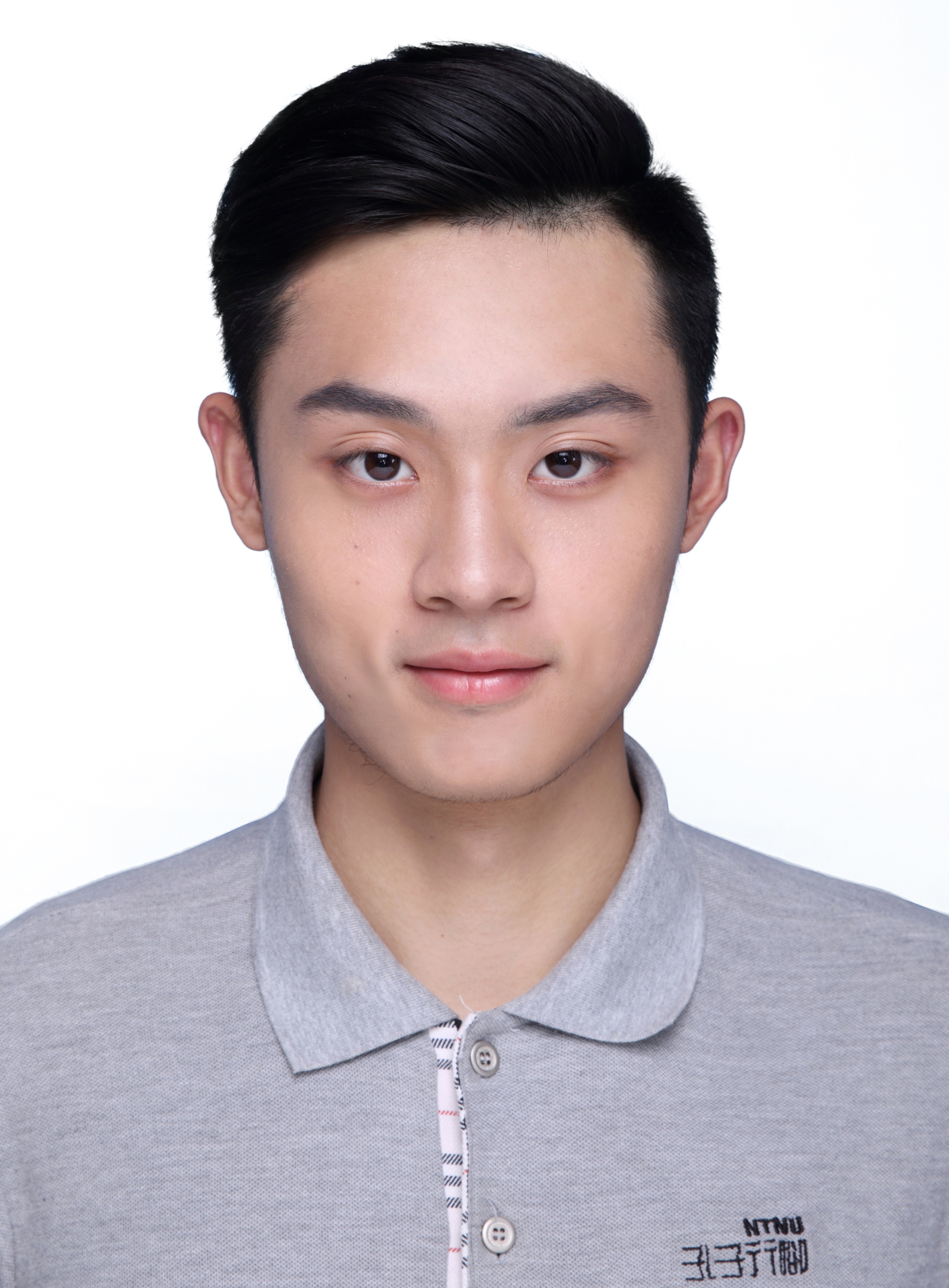}}]{Yifan Zhang}
is working toward the M.E. degree in the School of Software Engineering, South China University of Technology, China. He received the B.E. degree from the Southwest University, China, in 2017. His research interests are broadly in machine learning and data mining.  He has published papers in top venues including SIGKDD, NeurIPS, TKDE, TIP, etc.  He has been invited as a PC member or reviewer for international conferences and journals, such as NeurIPS, MICCAI and NeuroComputing.
\end{IEEEbiography}

\begin{IEEEbiography}[{\includegraphics[width=1.2in,height=1.25in,clip,keepaspectratio]{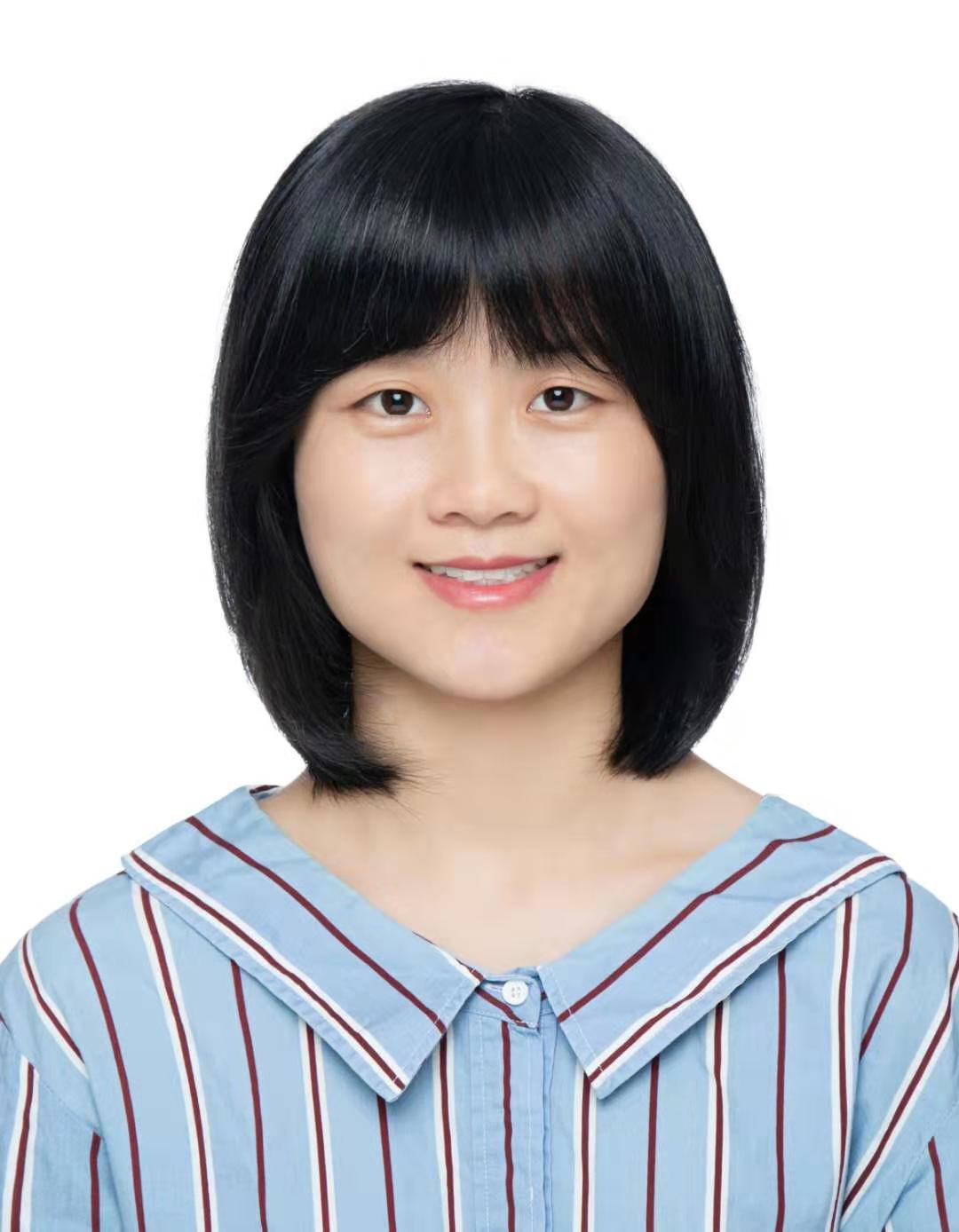}}]{Ying Wei} is a Senior Researcher at Tencent AI Lab. She works on machine learning, and is especially interested in solving challenges in transfer and meta learning by pushing the boundaries of both theories and applications. She received her Ph.D. degree from Department of Computer Science and Engineering, Hong Kong University of Science and Technology in 2017. Before this, she completed her Bachelor degree from Huazhong University of Science and Technology in 2012, with first class honors. She has published her work in top venues including ICML, NeurIPS, AAAI, and etc. She has been invited as a senior PC member for AAAI, and a PC member for many other top-tier conferences, such as ICML, NeurIPS, ICLR, and etc.
\end{IEEEbiography}

\begin{IEEEbiography}[{\includegraphics[width=1.2in,height=1.2in,clip,keepaspectratio]{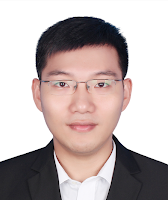}}]{Qingyao Wu}
received the Ph.D. degree in computer science from the Harbin Institute of Technology,
Harbin, China, in 2013. He was a Post-Doctoral Research Fellow with the School of Computer Engineering, Nanyang Technological University, Singapore, from 2014 to 2015. He is currently a Professor with the School of Software Engineering, South China University of Technology, Guangzhou, China. His current research interests include machine learning, data mining, big data research.
\end{IEEEbiography}

\begin{IEEEbiography}[{\includegraphics[width=1.2in,height=1.25in,clip,keepaspectratio]{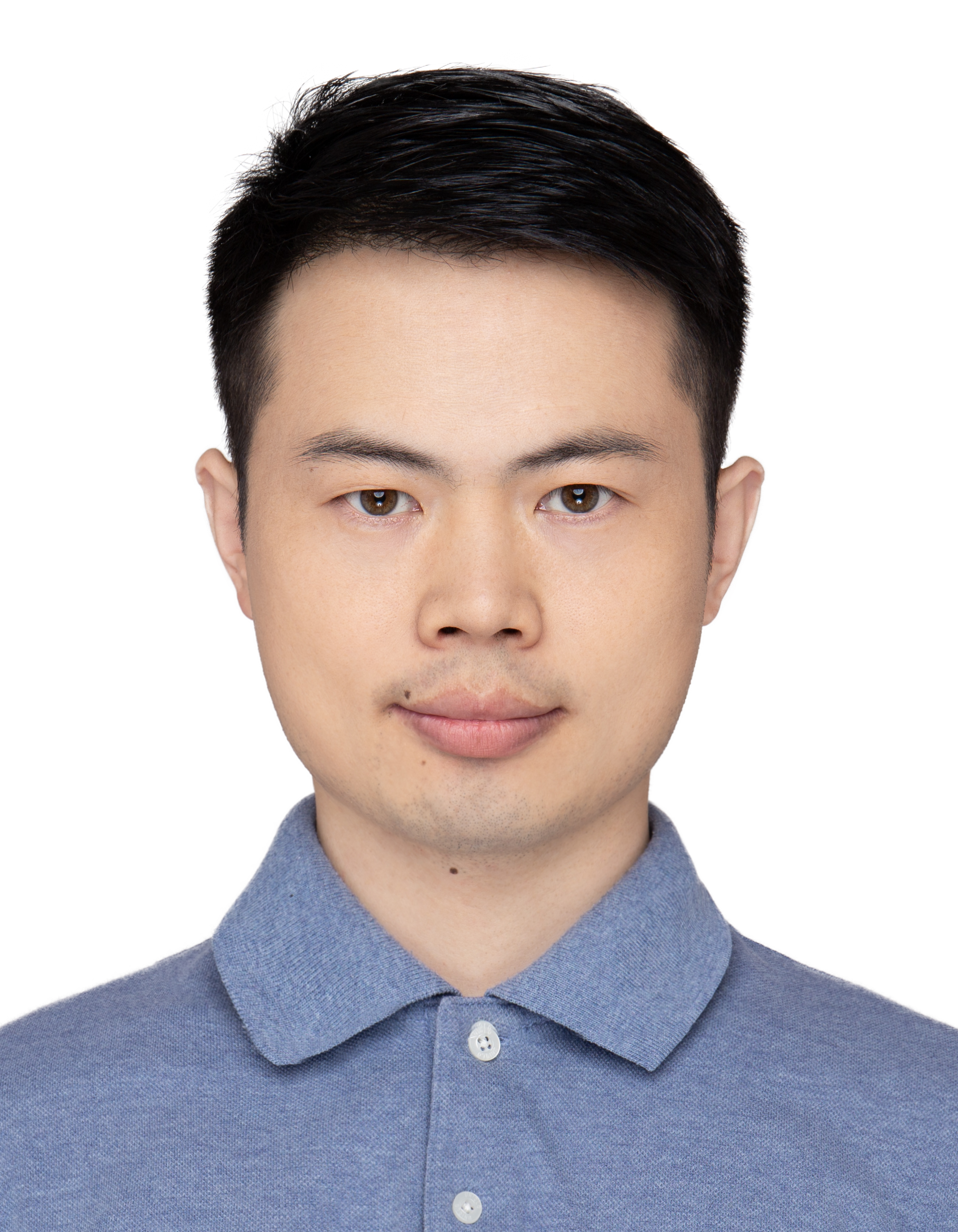}}]{Peilin Zhao}
is currently a Principal Researcher at Tencent AI Lab, China. Previously, he has worked at Rutgers University, Institute for Infocomm Research (I2R), Ant Financial Services Group. His research interests include: Online Learning, Recommendation System, Automatic Machine Learning, Deep Graph Learning, and Reinforcement Learning etc. He has published over 100 papers in top venues, including JMLR, ICML, KDD, etc. He has been invited as a PC member, reviewer or editor for many international conferences and journals, such as ICML, JMLR, etc. He received his bachelor’s degree from Zhejiang University, and his Ph.D. degree from Nanyang Technological University.
\end{IEEEbiography}

\begin{IEEEbiography}[{\includegraphics[width=1.2in,height=1.25in,clip,keepaspectratio]{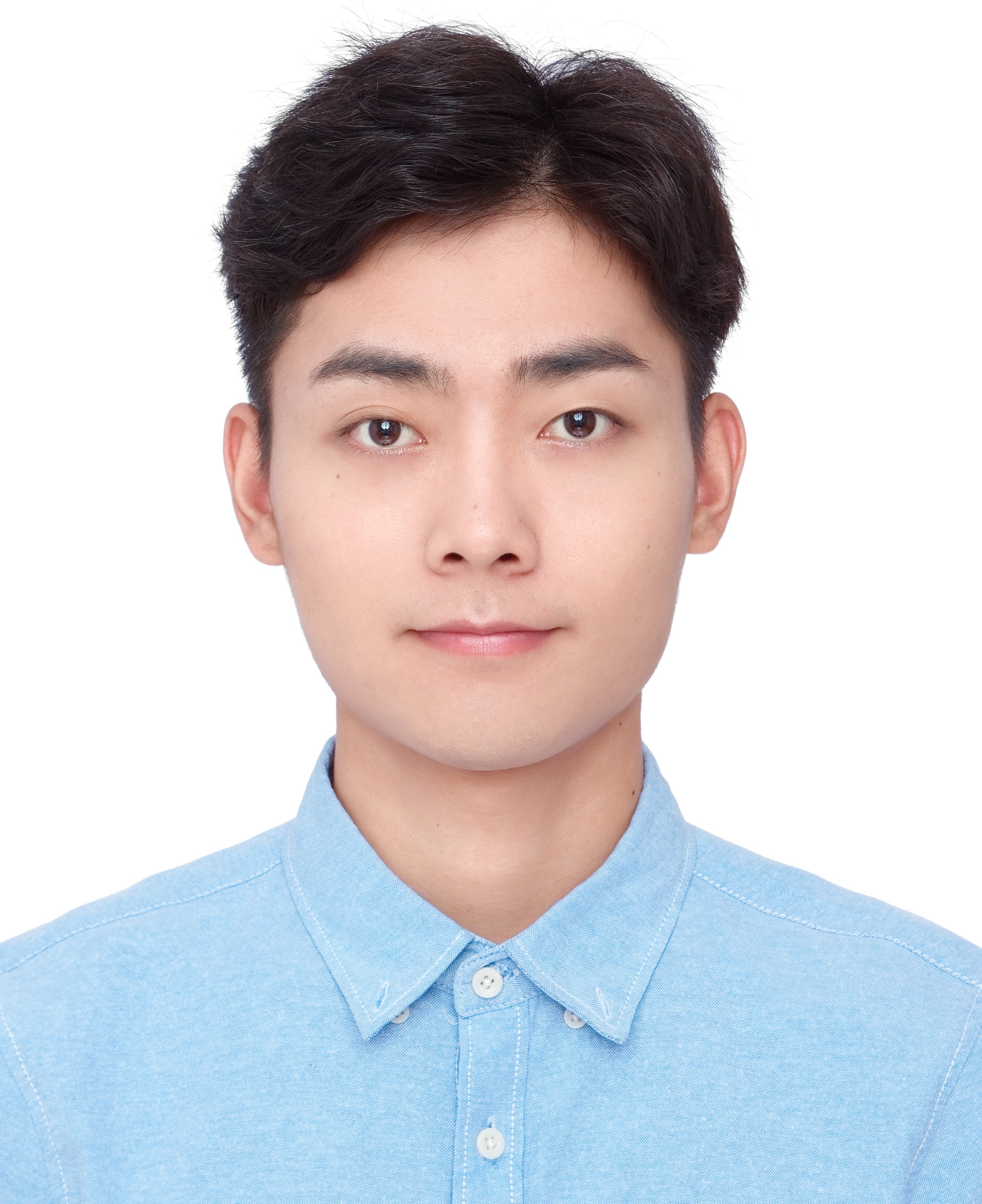}}]{Shuaicheng Niu}
is currently a Ph.D. candidate in South China University of Technology (SCUT), China, School of Software Engineering. He received the Bachelor's degree in mathematics from Southwest Jiaotong University (SWJTU), China, in 2018. His research interests include Machine Learning, Reinforcement Learning and Neural Network Architecture Search.
\end{IEEEbiography}

\begin{IEEEbiography}[{\includegraphics[width=1in,height=1.25in,clip,keepaspectratio]{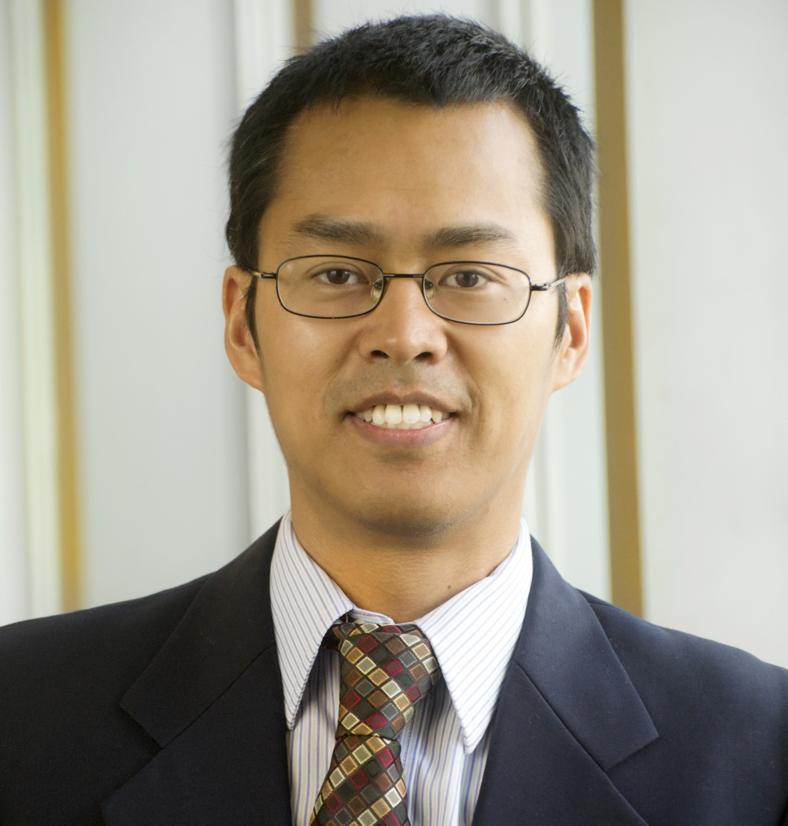}}]{Junzhou Huang}
is an Associate Professor in the Computer Science and Engineering department at the University of Texas at Arlington. He received the B.E. degree from Huazhong University of Science and Technology, China, the M.S. degree from Chinese Academy of Sciences, China, and the Ph.D. degree in Rutgers University. His major research interests include machine learning, computer vision and imaging informatics. He was selected as one of the 10 emerging leaders in multimedia and signal processing by the IBM T.J. Watson Research Center in 2010. He received the NSF CAREER Award in 2016.
\end{IEEEbiography}

\begin{IEEEbiography}[{\includegraphics[width=1.2in,height=1.25in,clip,keepaspectratio]{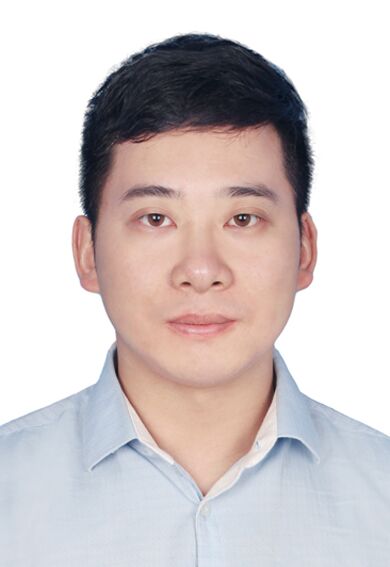}}]{Mingkui Tan}
received his Bachelor's Degree in Environmental Science and Engineering in 2006 and Master's degree in Control Science and Engineering in 2009, both from Hunan University in Changsha, China. He received the Ph.D. degree in Computer Science from Nanyang Technological University, Singapore, in 2014.  From 2014-2016, he  worked as a Senior Research Associate on computer vision in the School of Computer Science, University of Adelaide, Australia.  Since 2016, he has been with the School of Software Engineering, South China University of Technology, China, where he is currently a Professor.   His research interests include  machine learning,  sparse analysis, deep learning and large-scale optimization.
\end{IEEEbiography}
\end{document}


\title{Supplementary Materials}

\author{Yifan Zhang, Ying Wei, Qingyao Wu, Peilin Zhao, Shuaicheng Niu, Junzhou Huang, Mingkui Tan
\thanks{Y. Zhang,  S. Niu are with  South China University of Technology, and Pazhou Lab. E-mail: $\{$sezyifan, sensc$\}$@mail.scut.edu.cn.}
\thanks{Y. Wei, P. Zhao and J. Huang are with Tencent AI Lab, China. Email: $\{$judywei, masonzhao, joehhuang$\}$@tencent.com.}
\thanks{Q. Wu and M. Tan are with South China University of Technology, China. E-mail: $\{$qyw, mingkuitan$\}$@scut.edu.cn.}

}
\markboth{IEEE Transactions on Image Processing}
{Shell \MakeLowercase{\textit{et al.}}: Bare Demo of IEEEtran.cls for IEEE Journals}
\maketitle

\begin{abstract}
This supplementary material provides the proof for all theorems, architecture details and additional experimental results  in our paper ``Collaborative Unsupervised Domain Adaptation for Medical Image Diagnosis”.
\end{abstract}

\begin{IEEEkeywords}
Unsupervised Domain Adaptation, Deep Learning, Noisy Label, Medical Image Diagnosis
\end{IEEEkeywords}

\IEEEpeerreviewmaketitle

\section{Definition and Proof for Theorems}

\subsection{Definitions}

Following~\cite{Chazelle2001the,Mansour2009domain}, we define the discrepancy distance between the source distribution $\mathcal{S}$ and the target distribution $\mathcal{T}$ as:
$ {\rm disc}_{\mathcal{L}}(\mathcal{S},\mathcal{T})\small{=}\max_{h_1,h_2\in H}\big{|} \mathcal{L}_{\mathcal{S}}(h_1,h_2)\small{-}\mathcal{L}_{\mathcal{T}}(h_1,h_2)\big{|},$ where $H$ means a set of hypothesis, and $\mathcal{L}$ denotes some loss function. Since our analysis based on the Rademacher complexity, we then give its definition and empirical counterpart.

\begin{definition} {\rm \textbf{(Rademacher Complexity)}}
Let $H$ be a set of real-valued functions. The empirical Rademacher complexity of $H$ over data distribution $\mathcal{X}$ is defined as:
\begin{align}
 \mathfrak{\hat{R}}_{\mathcal{X}}(H)=\frac{2}{m}\mathbb{E}_{\sigma}\Big{[}\sup_{h\in H}|\sum_{i=1}^m\sigma_i h(x_i)|\Big{|}x_1,x_2,...,x_m \in \mathcal{X}\Big{]},\nonumber
\end{align}
where $\sigma_1, \sigma_2,...,\sigma_m$ are independent uniform $\{\pm 1\}$-valued random variables. Then, the Rademacher complexity of $H$ is $\mathfrak{R}_{\mathcal{X}}(H)=\mathbb{E} \mathfrak{\hat{R}}_{\mathcal{X}}(H)$.
\end{definition}

\subsection{Proof of Theorem 1}
\begin{theorem}\label{theorem}
Let $U$ be a hypothesis set for the domain loss $\mathcal{L}^{d}$ and $H$ be a hypothesis set for the classification loss $\mathcal{L}^{c}\in [0,M]$ in terms of $K$ classes. Assume that the loss function $\mathcal{L}$ is symmetric and obeys the triangle inequality. Suppose the noise transition matrix $Q$ is invertible and known.  For any $\delta>0$ and any hypothesis $h\in H$, with probability at least $1-(2\small{+}K)\delta$ over $m$ samples drawn from $\mathcal{S}$ and $n$ samples drawn from $\mathcal{T}$,  the following holds:
\begin{align}
  &\mathcal{L}_{\mathcal{T}}^{c}(\hat{h}^*_{\mathcal{\hat{S}'}},l_{\mathcal{T}}) -  \mathcal{L}_{\mathcal{T}}^{c}(h^*_{\mathcal{T}},l_{\mathcal{T}}) \nonumber \\
  & \leq {\rm disc}_{\mathcal{L}^{d}}(\mathcal{\hat{S}},\mathcal{\hat{T}})  +\mathcal{L}_{\mathcal{S}}^{c}(h^*_{\mathcal{S}},h^*_{\mathcal{T}})
  + 4K\mathfrak{\hat{R}}_{\mathcal{S}}(H) \nonumber \\
   &+8 \mathfrak{\hat{R}}_{\mathcal{S}}(U)+8\mathfrak{\hat{R}}_{\mathcal{T}}(U) \small{+}9\Bigg{(}(1\small{+}\frac{2}{3}M)\sqrt{\frac{\log \frac{2}{\delta}}{2m}} \small{+}\sqrt{\frac{\log \frac{2}{\delta}}{2n}}\Bigg{)}.\nonumber
\end{align}
\end{theorem}

\begin{proof}
{In CoUDA, two peer networks share the same architecture and optimization procedure, so two networks have the same generalization bound. For clarity, we first analyze a single one.
To begin with, relying on Theorem 8 in~\cite{Mansour2009domain}, we have the following proposition.}
\begin{prop}\label{prop1}
{Assume that the loss function $\mathcal{L}$ is symmetric and obeys the triangle inequality. For any hypothesis $h\in H$, the following holds:}
\begin{align}
  \mathcal{L}_{\mathcal{T}}^{c}(h,l_{\mathcal{T}}) \leq & \mathcal{L}_{\mathcal{T}}^{c}(h^*_{\mathcal{T}},l_{\mathcal{T}})+\mathcal{L}_{\mathcal{S}}^{c}(h,h^*_{\mathcal{S}}) \nonumber \\ &+{\rm disc}_{\mathcal{L}^{d}}(\mathcal{S},\mathcal{T})+\mathcal{L}_{\mathcal{S}}^{c}(h^*_{\mathcal{S}},h^*_{\mathcal{T}}).\nonumber
\end{align}
\end{prop}

{From Proposition~\ref{prop1},} the hypothesis error regarding to the target domain $\mathcal{L}_{\mathcal{T}}^{c}(h,l_{\mathcal{T}}) - \mathcal{L}_{\mathcal{T}}^{c}(h^*_{\mathcal{T}},l_{\mathcal{T}})$ is related with the average source classification loss $\mathcal{L}_{\mathcal{S}}^{c}(h,h^*_{\mathcal{S}})$, the domain discrepancy ${\rm disc}_{\mathcal{L}^{d}}(\mathcal{S},\mathcal{T})$ and the average loss between the best intra-class hypotheses $\mathcal{L}_{\mathcal{S}}^{c}(h^*_{\mathcal{S}},h^*_{\mathcal{T}})$. For adaptation to succeed, it is natural to assume that $\mathcal{L}_{\mathcal{S}}^{c}(h^*_{\mathcal{S}},h^*_{\mathcal{T}})$ is small~\cite{Mansour2009domain}, where more discussion is put in Section C in the main text.
Next, we mainly discuss the remaining two terms, and let us begin with the domain discrepancy ${\rm disc}_{\mathcal{L}^{d}}(\mathcal{S},\mathcal{T})$.
\begin{lemma} \label{lemma1}
Let $U$ be a hypothesis set for the domain loss $\mathcal{L}^{d}$. Then, for any $\delta>0$, with probability at least $1-2\delta$ over $m$ samples drawn according to $\mathcal{S}$ and $n$ samples drawn according to $\mathcal{T}$, we have:
\begin{align}
  {\rm disc}_{\mathcal{L}^{d}}&(\mathcal{S},\mathcal{T})  \leq    {\rm disc}_{\mathcal{L}^{d}}(\mathcal{\hat{S}},\mathcal{\hat{T}}) + \nonumber \\
  & 8 \Big{(}\mathfrak{\hat{R}}_{\mathcal{S}}(U)+\mathfrak{\hat{R}}_{\mathcal{T}}(U)\Big{)} +9\Bigg{(}\sqrt{\frac{\log \frac{2}{\delta}}{2m}}+\sqrt{\frac{\log \frac{2}{\delta}}{2n}}\Bigg{)}.\nonumber
\end{align}
\end{lemma}
Here, $\mathfrak{\hat{R}}(\cdot)$ is the empirical Rademacher complexity. See  Appendix A.3 for the proof.
This lemma shows that the discreapncy distance between domains can be estimated from finite samples. Next, we analyze the performance in the source domain with noisy labels.

\begin{lemma}\label{lemma2}
Assume $H$ be a hypothesis set for the modified focal loss $\mathcal{L}^{c}\in [0,M]$ in terms of $K$-class classification and suppose the noise transition matrix Q is invertible and known.  Then, for any $\delta>0$, with probability at least $1\small{-}K\delta$ over $m$ samples drawn according to $\mathcal{S}$, the following holds:
\begin{align}
  R_{\mathcal{S}}(\hat{h}^*_{\mathcal{\hat{S}'}})\leq R_{\mathcal{S}}(h^*_{\mathcal{S}})+4K\mathfrak{\hat{R}}_{\mathcal{S}}(H)+6M\Bigg{(}\sqrt{\frac{\log \frac{2}{\delta}}{2m}}\Bigg{)}.\nonumber
\end{align}
\end{lemma}

See  Appendix A.4 for the proof. This lemma shows that the optimal classifier $h^*_{\mathcal{S}}$ learned with true labels can be estimated from finite samples by a good classifier $\hat{h}^*_{\mathcal{\hat{S}'}}$ learned with noisy labels.

{Since the prediction of the collaborative network is a convex combination of the predictions of both networks, according to ~\cite{ref_Rademacher_Bartlett,Koltchinskii2002empirical}, the generalization bound of the collaborative networks is the same to that of a single peer network. Therefore, based on Proposition 1, Lemma Sup.~\ref{lemma1} and Lemma Sup.~\ref{lemma2}, we conclude the proof of Theorem 1.}
\end{proof}

\subsection{Proof of Lemma Sup.~\ref{lemma1}}
\begin{proof}
The following inequality is a version of the Rademacher complexity bound~\cite{ref_Rademacher_Bartlett}.
Let $H$ be a class of functions mapping $X,Y$ to $[0,M]$, and there are $m$ samples drawn according to a distribution $\mathcal{X}$. Then, for any $\delta>0$, with probability at least $1-\delta$ over these samples, the following holds for all $h\in H$:
\begin{align}\label{rademacher_bound}
      R_{\mathcal{X}}(h) \leq \hat{R}_{\mathcal{X}}(h)+\mathfrak{\hat{R}}_{\mathcal{X}}(H)+3M\sqrt{\frac{\log \frac{2}{\delta}}{2m}}.
\end{align}
The proof method can be found in~\cite{ref_Rademacher_Bartlett}.

For a bounded loss function $\mathcal{L}$, we can bound the discrepancy distance between a distribution and its empirical distribution, based on Rademacher complexity of the function set $\mathcal{L}_H=\{x\rightarrow |h_1(x)-h_2(x)|^q: h_1,  h_2 \in H\}$. Assume the loss function $\mathcal{L}$ is bounded by $M\geq 0$. Let $\mathcal{X}$ be a distribution and $\mathcal{\hat{X}}$ be the corresponding empirical distribution. Then, for any $\delta>0$, with probability at least $1-\delta$ over $m$ samples drawn according to $\mathcal{X}$, the following holds for all $h\in H$:
\begin{align}\label{discrepancy_bound}
  {\rm disc}_{\mathcal{L}}(\mathcal{X},\mathcal{\hat{X}}) \leq \mathfrak{\hat{R}}_{\mathcal{X}}(\mathcal{L}_H)+3M\sqrt{\frac{\log \frac{2}{\delta}}{2m}}.
\end{align}
The proof method can be found in~\cite{Mansour2009domain}.

In our proposed method, the domain loss $\mathcal{L}^{d}$ relies on the least square loss, \ie $L_2$ regression loss, which is 2-Lipschitz. Thus, by the contraction lemma~\cite{Ledoux2013}, we have:
\begin{align}\label{bound_L_H}
    \mathfrak{\hat{R}}_{\mathcal{X}}(\mathcal{L}_H^{d})\leq 4\mathfrak{\hat{R}}_{\mathcal{X}}(H')
\end{align}
with $H'=\{x\rightarrow (h_1(x)-h_2(x)):h_1,h_2\in H\}$.
Then, Based on the definition of the Rademacher variables and the sub-additivity of the supremum function, $\mathfrak{\hat{R}}_{\mathcal{X}}(H')$ can be bounded as follows:
\begin{align}\label{bound_H}
   &\mathfrak{\hat{R}}_{\mathcal{X}}(H')  \nonumber \\ &=\frac{2}{m}\mathbb{E}_{\sigma}\Big{[}\sup_{h_1,h_2\in H}|\sum_{i=1}^m\sigma_i (h_1(x_i)-h_2(x_i))|\Big{]}  \nonumber \\
  & \leq \frac{2}{m}\mathbb{E}_{\sigma}\Big{[}\sup_{h_1\in H}|\sum_{i=1}^m\sigma_i h_1(x_i)|\Big{]} \small{+}\frac{2}{m}\mathbb{E}_{\sigma}\Big{[}\sup_{h_2\in H}|\sum_{i=1}^m\sigma_i h_2(x_i)|\Big{]} \nonumber \\
  & = 2 \mathfrak{\hat{R}}_{\mathcal{X}}(H).
\end{align}

Moreover, the domain logits of the discriminator belong to $[0,1]$. With the transferability-aware weight $w\in[1,3]$, we can easily find that $\mathcal{L}^{d}$ is bounded by 3. Combining Eqs.~(\ref{discrepancy_bound}-\ref{bound_H}), for any $\delta>0$, with probability at least $1-\delta$ over $m$ samples drawn according to $\mathcal{X}$, the following holds for all $h\in H$:
\begin{align}\label{discrepancy_bound_new}
  {\rm disc}_{\mathcal{L}^{d}}(\mathcal{X},\mathcal{\hat{X}}) \leq 8\mathfrak{\hat{R}}_{\mathcal{X}}(H)+9\sqrt{\frac{\log \frac{2}{\delta}}{2m}}.
\end{align}

Then, by the triangle inequality, we have:
\begin{align}\label{triangle_bound}
   {\rm disc}&_{\mathcal{L}^{d}}(\mathcal{S},\mathcal{T}) \leq\nonumber \\
  & {\rm disc}_{\mathcal{L}^{d}}(\mathcal{S},\mathcal{\hat{S}}) +{\rm disc}_{\mathcal{L}^{d}}(\mathcal{\hat{S}},\mathcal{\hat{T}})  +{\rm disc}_{\mathcal{L}^{d}}(\mathcal{\hat{T}},\mathcal{T}).
\end{align}

Combining Eqs.~(\ref{discrepancy_bound_new}-\ref{triangle_bound}) and adjusting to the hypothesis $U$ for discriminator, we conclude the proof.
\end{proof}

\subsection{Proof of Lemma Sup.~\ref{lemma2}}
\begin{proof}
In our proposed method, we use a noise co-adaptation layer to approximate the noise transition matrix $Q$, so that the classifier learned with noisy labels is consistent with the optimal classifier learned with true labels.
To facilitate the analysis, following~\cite{Yu2018learning}, we assume the noise transition matrix $Q$ is known and invertible.
Besides, consider that the focal loss is a modified cross-entropy loss. According to Lemma 1 and Theorem 1 in~\cite{Yu2018learning}, the minimizer $h^*_{\mathcal{S'}}$ of $R_{\mathcal{S'}}(h)$ is also the minimizer $h^*_{\mathcal{S}}$ of $R_{\mathcal{S}}(h)$, \ie $h^*_{\mathcal{S'}}=h^*_{\mathcal{S}}$. Moreover, we know that, given sufficient training data with noisy labels, the empirical optimal $\hat{h}^*_{\mathcal{\hat{S'}}}$ converges to the expected optimal $h^*_{\mathcal{S'}}$~\cite{Natarajan2013learning}. Therefore,   $\hat{h}^*_{\mathcal{\hat{S'}}}$ can also converge to $h^*_{\mathcal{S}}$ even though the former is learned with noisy labels. Like~\cite{Natarajan2013learning,Yu2018learning}, we have:
\begin{align} \label{generalization_relation}
   & R_{\mathcal{S}}(\hat{h}^*_{\mathcal{\hat{S'}}})-R_{\mathcal{S}}(h^*_{\mathcal{S}})\nonumber \\
  &= R_{\mathcal{S'}}(\hat{h}^*_{\mathcal{\hat{S'}}})-R_{\mathcal{S'}}(h^*_{\mathcal{S}}) \nonumber \\
  & =\hat{R}_{\mathcal{\hat{S}'}}(\hat{h}^*_{\mathcal{\hat{S'}}})-\hat{R}_{\mathcal{\hat{S}'}}(h^*_{\mathcal{S}}) +R_{\mathcal{S'}}(\hat{h}^*_{\mathcal{\hat{S'}}})-\hat{R}_{\mathcal{\hat{S}'}}(\hat{h}^*_{\mathcal{\hat{S'}}})\nonumber \\ & ~~~~ +\hat{R}_{\mathcal{\hat{S}'}}(h^*_{\mathcal{S}})  -R_{\mathcal{S'}}(h^*_{\mathcal{S}}) \nonumber \\
  & \leq R_{\mathcal{S'}}(\hat{h}^*_{\mathcal{\hat{S'}}})-\hat{R}_{\mathcal{\hat{S}'}}(\hat{h}^*_{\mathcal{\hat{S'}}})
  +\hat{R}_{\mathcal{\hat{S}'}}(h^*_{\mathcal{S}})  -R_{\mathcal{S'}}(h^*_{\mathcal{S}}) \nonumber \\
  & \leq 2 \sup_{h\in H}| R_{\mathcal{S'}}(h)-\hat{R}_{\mathcal{\hat{S}'}}(h)|,
\end{align}
since $\hat{R}_{\mathcal{\hat{S}'}}(\hat{h}^*_{\mathcal{\hat{S'}}})-\hat{R}_{\mathcal{\hat{S}'}}(h^*_{\mathcal{S}})\leq 0$. Moreover, the error in the last line is called the generalization error.
To analyze the generalization error for multi-class classification,  inspired by ~\cite{Wan2013Regularization,Zhai2018}, we first reformulate the focal loss on the top of the softmax into a single modified logistic loss function
$  -\sum_{i=1}^K z^i (1-o^i)\log\frac{\exp o^i}{\sum_j \exp o^j }$, where $z^i$ is the noise label in terms of the $i$-th class,  $o^i$ is the corresponding prediction logits, and $K$ is the class number.

Then, relying on the concentration lemma~\cite{Ledoux2013} and  Lemma 3 in~\cite{Wan2013Regularization}, and considering $1-o^i\leq 1$, the generalization error for multi-class classification can be upper bounded with Rademacher complexity~\cite{ref_Rademacher_Bartlett} as follows.  For any $\delta>0$, with probability at least $1-K\delta$ over $m$ samples drawn according to $\mathcal{S'}$, we have:
\begin{align}\label{generalization_bound}
    \sup_{h\in H}| R_{\mathcal{S'}}(h)-\hat{R}_{\mathcal{\hat{S}'}}(h)|\leq   2K\mathfrak{\hat{R}}_{\mathcal{S}}(H)+3M\Bigg{(}\sqrt{\frac{\log \frac{2}{\delta}}{2m}}\Bigg{)},
\end{align}
where $M$ is the upper bound of the single modified logistic loss. Combining Eqs. (\ref{generalization_relation}-\ref{generalization_bound}), we conclude the proof of Lemma Sup.~\ref{lemma2}.
\end{proof}

\begin{table*}[t]
	\caption{Detailed architecture of the base network in the cooperative network, where peer networks share the same architecture. Here, $m$ and $k$ indicate the length and width of the input image and K means the class number, while GAP represents global average pooling. All spatial convolutions use 3x3 kernels except  conv2d 1x1, which uses 1x1 kernels.}
	\vspace{-0.05in}
	\label{table:architecture}
	\centering
	\renewcommand*\arraystretch{1.4}
		\begin{tabular}{c|c|c|c|c|c}
			\hline
			\hline
			\multicolumn{6}{c}{\textbf{Backbone Network}} \\
			\hline
			\hline
			{Part} & {Input $ \rightarrow $ Output} & {Expansion}& {Repeat}& {Stride}& {Activation}\\
			\hline
			\multirow{1}[0]{*}{conv2d}
			& $ (m, k, 3) \rightarrow (\frac{m}{2}, \frac{k}{2}, 32) $ & -&1&2& ReLU  \\
			\hline
			\multirow{1}[0]{*}{bottleneck}
			& $ (\frac{m}{2}, \frac{k}{2}, 32) \rightarrow (\frac{m}{2}, \frac{k}{2}, 16) $ &1&1&1 &ReLU  \\
			\hline
			\multirow{1}[0]{*}{bottleneck}
			& $ (\frac{m}{2}, \frac{k}{2}, 16) \rightarrow (\frac{m}{4}, \frac{k}{4}, 24) $   & 6&2&2 &ReLU  \\
			\hline
			\multirow{1}[0]{*}{bottleneck}
			& $ (\frac{m}{4}, \frac{k}{4}, 24) \rightarrow (\frac{m}{8}, \frac{k}{8}, 32) $   & 6&3&2 &ReLU  \\
			\hline
			\multirow{1}[0]{*}{bottleneck}
			& $ (\frac{m}{8}, \frac{k}{8}, 32) \rightarrow (\frac{m}{16}, \frac{k}{16}, 64) $  & 6&4&2 &ReLU  \\
			\hline
			\multirow{1}[0]{*}{bottleneck}
			& $ (\frac{m}{16}, \frac{k}{16}, 64) \rightarrow (\frac{m}{16}, \frac{k}{16}, 96) $   & 6&3&1 &ReLU  \\
			\hline
			\multirow{1}[0]{*}{bottleneck}
			& $ (\frac{m}{16}, \frac{k}{16}, 96) \rightarrow (\frac{m}{32}, \frac{k}{32}, 160) $ & 6&3&2 &ReLU  \\
			\hline
			\multirow{1}[0]{*}{conv2d 1x1}
			& $ (\frac{m}{32}, \frac{k}{32}, 160) \rightarrow (\frac{m}{32}, \frac{k}{32}, 320) $ & -&1&1 &ReLU  \\	
			\hline
			\multirow{1}[0]{*}{GAP}
			& $ (\frac{m}{32}, \frac{k}{32}, 320) \rightarrow (1, 1, 1280) $ & -&1&1 &ReLU  \\
			\hline
			\multirow{1}[0]{*}{conv2d 1x1}
			& $ (1, 1, 1280) \rightarrow (1, K) $ & -&1&- & Softmax  \\

			\hline
			\hline
			\multicolumn{6}{c}{\textbf{Noise  Co-adaptation Layer}} \\
            \hline
            \hline
			\multirow{1}[0]{*}{conv2d 1x1}
			& $ (1, 1, 1280) \& (1,K) \rightarrow (1,K) $ &  -  & K & 1 &   Softmax \\
			\hline
			\hline
		\end{tabular}
\end{table*}

\begin{table*}[ht]
	\caption{Discriminator network architecture.}
	\vspace{-0.05in}
	\label{table:discriminator}
	\centering
	\renewcommand*\arraystretch{1.4}
	\begin{tabular}{c|c|c}
			\hline
			\hline
			\multicolumn{3}{c}{\textbf{Discriminator}} \\

		\hline
		\hline
		{Layer} & {Input $ \rightarrow $ Output} & {Layer information} \\
		
		\hline
		Hidden Layer & $ (1, 1, 1280) \rightarrow (1,  1280) $ & Fully Connectly Layer, Leaky ReLU \\
		\hline
		Hidden Layer & $ (1,  1280) \rightarrow (1,  1280) $ & Fully Connectly Layer, Leaky ReLU \\
	    \hline
		Output  & $   (1,  1280) \rightarrow (1,  1) $ & Fully Connectly Layer, Sigmoid\\
		\hline
		\hline
	\end{tabular}
\end{table*}

\section{Architecture and More Discussions of CoUDA}
{This appendix details the architecture of the cooperative network.
Since two peer networks have the same network architecture, we here describe a single one. Specifically, it consists of three main components:  the backbone network (\ie the feature extractor, the classifier),  the noise co-adaptation layer, and the discriminator network.}

\textbf{Backbone network.}
To satisfy the resource and efficiency requirements of real-world medical tasks, we adopt  MobileNetV2~\cite{ref_Mobilenet} as the backbone.
The architecture is summarized in Table~\ref{table:architecture}.
We recommend readers to the paper~\cite{ref_Mobilenet} for more detailed information.

\textbf{Noise co-adaptation layer.}

{As shown in Table~\ref{table:architecture},
the  repeat number $K$ means that we have $K$ additional softmax layers. Each one serves for one class to estimate the transition probabilities from this class to pseudo-classes  (See Eq.~3 in the main text). After that, we can obtain the predictions of noisy labels based on the estimated noise transition probabilities (based on Eq.~4 in the main text).}

{In addition, a careful \textbf{initialization} of the parameters of the noise co-adaptation layer is crucial for successful convergence of the network into a good classifier at the test time. To this end, we follow the initialization strategy in~\cite{ref_noiselayer}, where we initialize all weight parameters as zero, and initialize the bias parameters by assuming a small uniform noise $\epsilon$ as:
\begin{align}\label{initial}
  b_{ij} = \log\Big{(}(1-\epsilon)\mathbb{I}_{(i=j)}+\frac{\epsilon}{K-1}\mathbb{I}_{(i\neq j)}\Big{)},
\end{align}  where the value of $\epsilon$ depends on the tasks in hand, and $i,j$ represent the true label and noisy label.}

However, the noise co-adaptation layer may suffer the \textbf{sclability} issue when the class number $K$ is very large, since there are $K$ additional softmax layers. In this case, the manner in Eq.~9 mentioned above can be a suitable choice to reduce the computational burden, since it does not have weight parameters. In fact, such a simplified method also performs well in the methods of the noise adaptation layer~\cite{ref_noiselayer,ref_Sukhbaatar2014}.

\textbf{Discriminator.} As for adversarial learning, inspired by ADDA~\cite{ref_ADDA}, we use three fully connected layers as the discriminator $D$, which is shown in Table~\ref{table:discriminator}.

\section{More Experimental Results}

\subsection{{Discussion about Transferability-aware Weights}}
{In the definition of the transferability-aware weight, we use the cosine distance to represent the distance similarity. In fact, one can also use other metrics like L1 or L2 distances. However, in our preliminary experiments, as shown in Table~\ref{measure}, there are no apparent differences among them, and the cosine distance performs slightly better. Therefore, we adopt the cosine distance in our method.}

\begin{table}[H]
	\caption{Comparisons of different distances for computing   transferability-aware weights on Colon-A.}
	\label{measure}
\begin{footnotesize}
 \begin{center}
   \begin{threeparttable}
 	\begin{tabular}{lcccc}\toprule
        Distances& Acc (\%)& MP &  MR  & Macro F1  \cr
        \midrule
         L1 distance  &87.27 	&86.68	& 86.75	& 86.72                    \\
          L2 distance &87.49	& 87.04	& \textbf{86.90}	& 86.97 \\

         cosine distance (ours) \ & \textbf{87.75} 	&  \textbf{87.62}	& 86.85 	&  \textbf{87.22}        \\
        \bottomrule
	\end{tabular}
	\end{threeparttable}
	 \end{center}
	  \end{footnotesize}
\end{table}

\subsection{{Discussion about Domain Loss}}
{In our method, we define the domain loss relying on the least square distance, since it helps to improve domain confusion and stabilize training, by preserving the domain distance information~\cite{ref_LSGAN}. In this section, we empirically compare it with the classic GAN loss~\cite{ref_GAN} with transferability-weights on Colon-A. As shown in Table~\ref{domain_measure}, the least-square loss performs better than GAN loss, which demonstrates the effectiveness of the adopted domain loss.}

\begin{table}[h]
	\caption{Evaluation the least-square distance as the domain loss on Colon-A.}\vspace{0.1in}
	\label{domain_measure}
\begin{footnotesize}
 \begin{center}
 \begin{threeparttable}
 	\begin{tabular}{lcccc}\toprule
        Domain Loss & Acc (\%)& MP &  MR  & Macro F1   \cr
        \midrule
           GAN loss  &86.45 	&85.40	& 86.17	& 85.77 \\
         least square loss (ours)   &  \textbf{87.75} 	&  \textbf{87.62}	& \textbf{86.85} 	&  \textbf{87.22}        \\
        \bottomrule
	\end{tabular}
	\end{threeparttable}
	 \end{center}
	  \end{footnotesize}
\end{table}

\subsection{{Discussion about Diversity Loss}}
{In our method, we define the diversity loss relying on the JS distance. In fact, one can also use other distance according to the tasks at hand. Nevertheless, we do not find apparent differences among these distances in our preliminary experiments, as shown in Table~\ref{discrepancy_measure}. Specifically, the JS distance performs relatively better.}

\begin{table}[H]
	\caption{Comparisons of different distances in the diversity loss on Colon-A.}
	\label{discrepancy_measure}
\begin{footnotesize}
 \begin{center}
 \begin{threeparttable}
 	\begin{tabular}{lcccc}\toprule
       Distances& Acc (\%)& MP &  MR  & Macro F1   \cr
        \midrule
          L1 distance  &87.32 	&86.99	& 87.16	& 87.07                    \\
          L2 distance &87.53	& 87.51	&86.69	& 87.07 \\
          KL divergence  &87.53 	&87.54	& 86.63 & 87.06                   \\
          cos distance &87.62	& 87.16	&  \textbf{87.24}	& 87.20 \\
          JS divergence (ours)   &  \textbf{87.75} 	&  \textbf{87.62}	& 86.85 	&  \textbf{87.22}        \\

        \bottomrule
	\end{tabular}
	\end{threeparttable}
	 \end{center}
	  \end{footnotesize}
\end{table}

\subsection{{Discussion about Best Intra-Class Hypotheses}}\label{entropy_hypothesis}
{This section discusses the average loss between the best intra-class hypotheses, \ie $\mathcal{L}_{\mathcal{S}}^{c}(h^*_{\mathcal{S}},h^*_{\mathcal{T}})$.
Previous work~\cite{Mansour2009domain}  assumes $\mathcal{L}_{\mathcal{S}}^{c}(h^*_{\mathcal{S}},h^*_{\mathcal{T}})$ be small, since if there is not any hypothesis that performs well on both domains, domain adaptation cannot be conducted. In fact, we can also minimize this term by enforcing entropy loss to guarantee intra-class consistency~\cite{ref_entropymin,ref_tcl}. Nevertheless, in our preliminary experiments as shown in Table~\ref{entropy}, entropy loss does not boost the performance a lot, so we forsake it for the simplicity of our proposed method.}
\begin{table}[H]
	\caption{Evaluation of the entropy loss on Colon-A.}
	\label{entropy}
\begin{footnotesize}
 \begin{center}
 \begin{threeparttable}
 	\begin{tabular}{lcccc}\toprule
        Methods& Acc (\%)& MP &  MR  & Macro F1   \cr
        \midrule
         CoUDA \small{+} entropy loss  &87.92 	&88.08	& 87.15	& 87.58                    \\
         CoUDA & 87.75 	& 87.62	& 86.85 	& 87.22        \\
        \bottomrule
	\end{tabular}
	\end{threeparttable}
	 \end{center}
	  \end{footnotesize}
\end{table}

\subsection{{Discussion about Ensemble Methods}}
{We adopt the average prediction of two peer networks as the final prediction, which is commonly used in  ensemble learning~\cite{ref_ensemble}. The promising experiments in the main text have demonstrated the effectiveness of the average ensemble.  In fact, one can also use the maximum ensemble, which even performs slightly better as shown in Table~\ref{ensemble}.  }

\begin{table}[H]
	\caption{Evaluation of different  ensemble methods on Colon-A.}
	\vspace{0.1in}
	\label{ensemble}
 \begin{center}
 \begin{threeparttable}
 	\begin{tabular}{ccccc}\toprule
        Methods  & Acc (\%) & MP & MR  & Macro F1  \cr
         \midrule
          Average Ensemble &87.75 	& 87.62	& 86.85	&87.22          \\
          Maximum Ensemble  & 88.14	& 87.58	& 88.20	&87.87    \\
        \bottomrule
	\end{tabular}
	 \end{threeparttable}
	 \end{center}\vspace{0.2in}
\end{table}

\subsection{{Discussion about the Number of Source Annotations}}
{We further evaluate the influence of the annotation number in the source domain, we test CoUDA with different numbers of WSIs on Colon-A. Table~\ref{number} shows that the more labeled WSIs, the better performance of CoUDA. Note that  with only 20,000 labeled WSIs,  CoUDA is able to perform well. }

\begin{table}[H]
\vspace{0.2in}
	\caption{Performance of CoUAD with different numbers of labeled WSIs on Colon-A.}
		\vspace{0.1in}
	\label{number}
 \begin{center}
 \begin{threeparttable}
 	\begin{tabular}{ccccc}\toprule
        Number of WSIs  & Acc (\%) & MP & MR  & Macro F1  \cr
         \midrule
          5,000 & 83.98 &82.70	& 81.41 	 	&  81.87         \\
          10,000 & 85.93	& 87.27 	& 82.25	& 83.73    \\
          20,000 & 87.23	& 87.28  & 85.14	& 86.05  \\
          44,542 & 87.75 	& 87.62	& 86.85	&87.22   \\
        \bottomrule
	\end{tabular}
	 \end{threeparttable}
	 \end{center}\vspace{0.2in}
\end{table}

\subsection{{Discussion about the Number of Peer Networks}}
{This section further evaluates  the influence of the number of peer networks on CoUDA. Table~\ref{peernetworks} shows the performance vs. the learning cost regarding different numbers of peer networks  on Colon-A.  To be specific, the increase in the number of peer networks improves the performance of CoUDA, while the magnitude of the increase gradually decreases. At the same time,  the number of  parameters and the  computational costs ($\#$FLOPs and
Memory consumption)  increase steadily. This may be a  huge concern in many real-world applications, \eg on mobile phones or intelligent microscopes. Therefore, it is better for users to select a suitable number of peer networks based on practical needs and computing resources.}

  \begin{table*}[t]
	\caption{Performance of CoUDA with different numbers of peer networks on Colon-A.}
	\label{peernetworks}
 \begin{center}
 \begin{threeparttable}
 	\begin{tabular}{cccccccc}\toprule
        $\#$Peer Networks  & Acc (\%) & MP & MR  & Macro F1 &  $\#$Params &  $\#$FLOPs & Memory   \cr
         \midrule
          2  &87.75 	& 87.62	& 86.85	&87.22  & 7.75M &  3.13G   &    16,783~MiB  \\
          3  & 90.26	& 91.55	& 87.99	&89.45  & 9.98M &  4.69G   &    23,421~MiB  \\
          4  & 92.08	& 91.82	& 92.07	& 91.91 & 12.20M &  6.26G   &    30,058~MiB  \\
          5  & 93.29	& 93.94	& 91.95	& 92.84 & 14.43M &  7.82G  &  36,697~MiB     \\
        \bottomrule
	\end{tabular}
	 \end{threeparttable}
	 \end{center}
\end{table*}

\subsection{{Visualization of Training Curve}}
\begin{figure*}[ht]
 \begin{minipage}{0.32\linewidth}
 \centerline{\includegraphics[width=5.5cm]{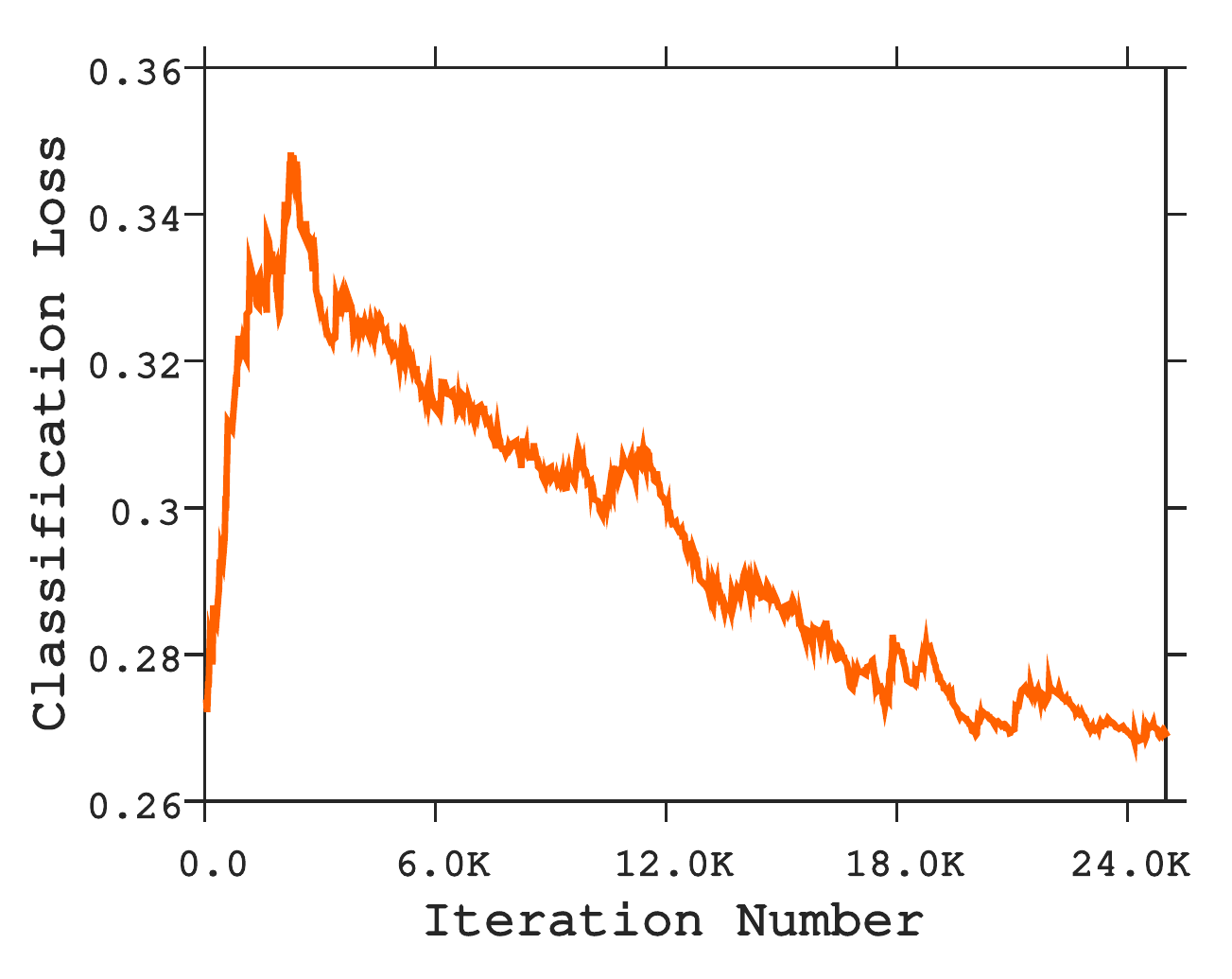}}\vspace{-0.075in}
 \centerline{(a)}
 \end{minipage}
 \hfill
  \begin{minipage}{0.32\linewidth}
 \centerline{\includegraphics[width=5.5cm]{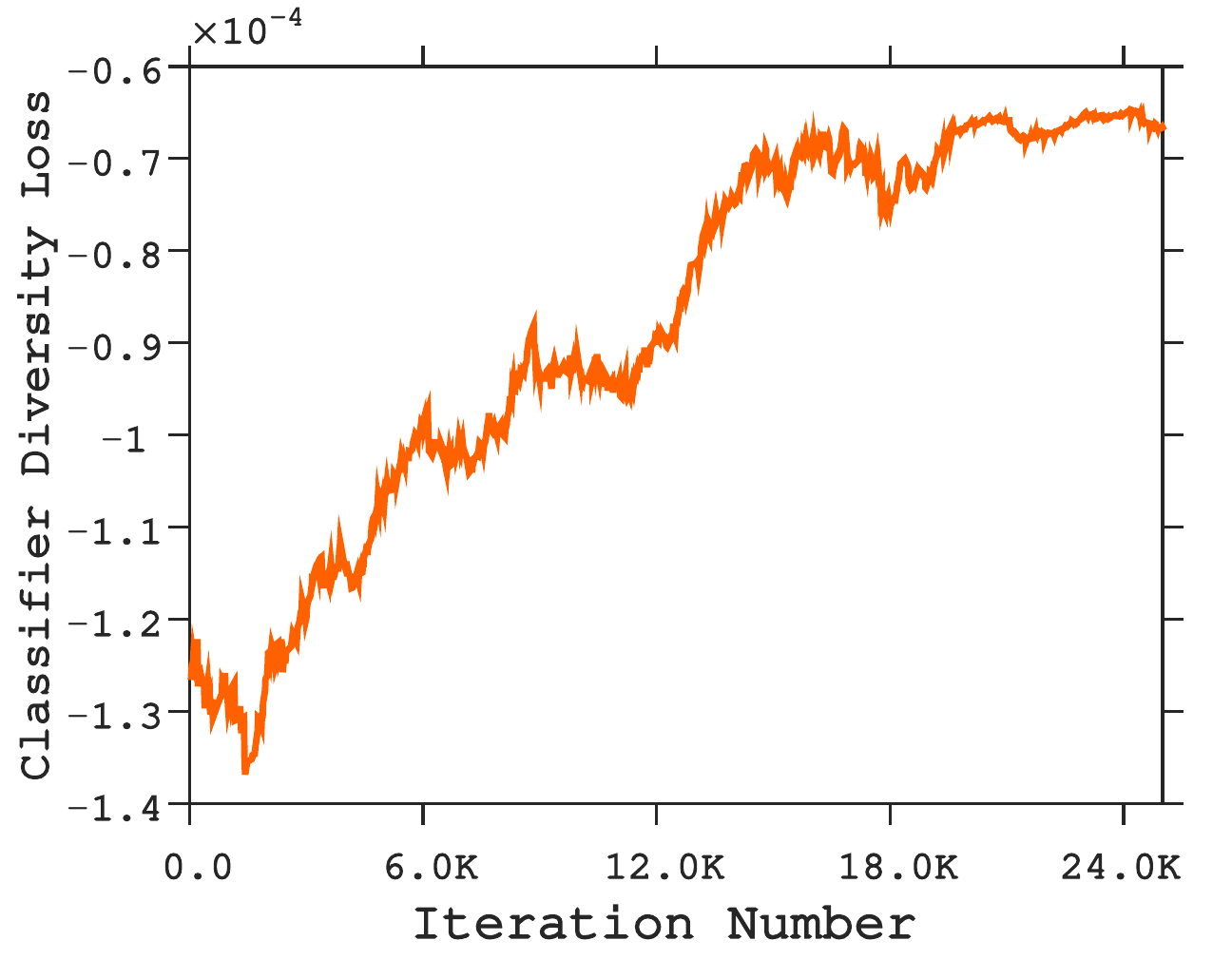}}\vspace{-0.075in}
 \centerline{(b)}
 \end{minipage}
  \begin{minipage}{0.32\linewidth}
 \centerline{\includegraphics[width=5.5cm]{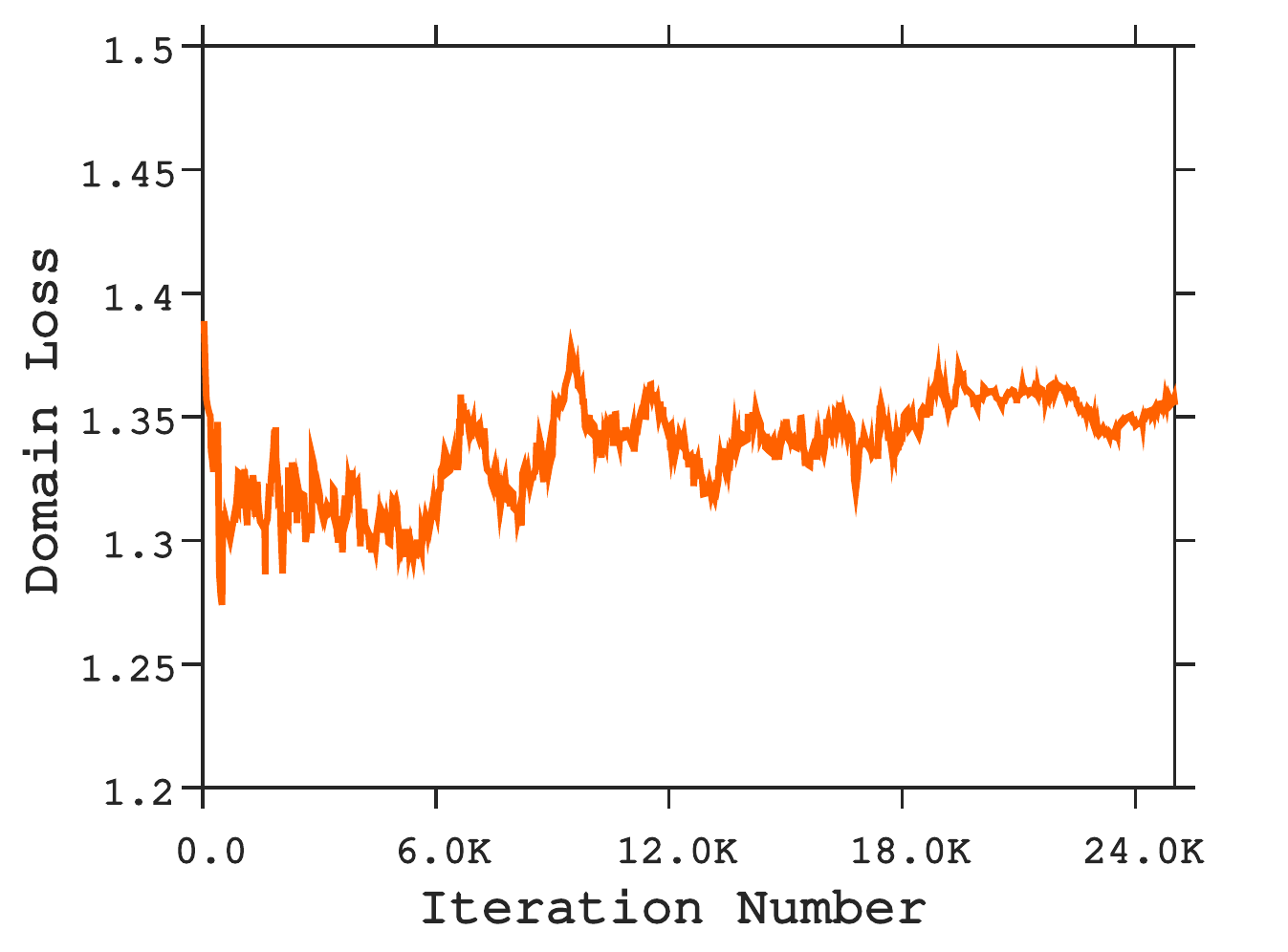}}\vspace{-0.075in}
 \centerline{(c)}
 \end{minipage}
 \caption{Training curves of CoUDA.}\label{loss_curves}
\end{figure*}

{In the proposed CoUDA method, by using the Gradient Reverse Layer (GRL), we can train all network components simultaneously  in an end-to-end manner. Note that GRL stabilises the adversarial training and  is widely used in adversarial domain adaptation~\cite{ref_DANN,zhang2019whole,zhang2019collaborative,ref_partial}. In fact, to improve the training efficiency, we first train the network backbone on the source WSI data, and then apply CoUDA to train the pre-trained model with both labeled WSIs and unlabeled MSIs. As for the latter process, we  show the training curve of each loss term in Fig.~\ref{loss_curves}.  To be specific, both the classification loss and the diversity loss evolve as expected, while the domain  loss gradually reaches equilibrium.}

 \newpage
 \balance
\bibliographystyle{IEEEtran}
\bibliography{CoUDA}